\PassOptionsToPackage{table,dvipsnames,xcdraw}{xcolor}
\documentclass[11pt]{article}

\usepackage[preprint]{acl}

\usepackage{times}
\usepackage{latexsym}

\usepackage[T1]{fontenc}

\usepackage[utf8]{inputenc}

\usepackage{microtype}

\usepackage{inconsolata}

\usepackage{graphicx}
\usepackage{dblfloatfix}
\usepackage{booktabs}
\usepackage{multirow}
\usepackage[normalem]{ulem}
\useunder{\uline}{\ul}{}
\usepackage{booktabs}
\usepackage{enumitem}
\usepackage{xcolor}  
\usepackage{tikz}
\usepackage[most]{tcolorbox}
\usepackage{graphicx}
\usepackage{pgfplots}
\pgfplotsset{compat=1.18}
\usepgfplotslibrary{groupplots}
\usepackage{filecontents}

\usepackage{amsthm}
\usepackage{amssymb}
\usepackage{subcaption}
\usepackage{colortbl}
\usepackage{pifont}
\usepackage{wrapfig}
\usepackage{booktabs}
\usepackage{array}
\usepackage{pifont}
\usepackage{makecell}
\usepackage{amsmath}
\usepackage{multirow}
\usepackage{algorithm}
\usepackage{algorithmic}
\usepackage{listings}
\usepackage{comment}

\lstdefinestyle{promptstyle}{
    frame=tb,
    basicstyle=\ttfamily\footnotesize,
    columns=fullflexible,
    breaklines=true,
    postbreak=\mbox{\textcolor{red}{$\hookrightarrow$}\space},
    captionpos=b,
    showstringspaces=false,
    numbers=none,
}

\usepackage{float}
\usepackage{siunitx}
\usepackage{enumitem}
\setlist[itemize]{leftmargin=*, noitemsep, topsep=0pt}

\definecolor{headerbg}{HTML}{EAF1F8}
\definecolor{ourbg}{HTML}{E8F5E9}
\definecolor{yesbg}{HTML}{D9EAD3}
\definecolor{partbg}{HTML}{FFF2CC}
\definecolor{nobg}{HTML}{F4CCCC}
\definecolor{lightrow}{HTML}{F8F9FA}

\definecolor{highgain}{HTML}{C8E6C9}
\definecolor{medgain}{HTML}{E8F5E9}
\definecolor{lowgain}{HTML}{FFF8E1}
\definecolor{best}{HTML}{A5D6A7}
\definecolor{reggain}{HTML}{FFCDD2}

\newcommand{\cmark}{\ding{51}}
\newcommand{\xmark}{\ding{55}}

\newcommand{\benchname}{\textsc{AFTER}}
\newcommand{\framework}{\textsc{Evolution}}

\renewcommand{\arraystretch}{0.95}
\usepackage[skip=2pt]{caption}

\title{Managing Procedural Memory in LLM Agents: \\ Control, Adaptation, and Evaluation}

\author{
	Julia Belikova \And
	Rauf Parchiev \And
	Evgeny Egorov \AND
	Grigorii Davydenko \And
	Gleb Gusev \And
	Andrey Savchenko \AND
	Maksim Makarenko
}

\begin{document}
\maketitle

\begin{abstract}


Procedural memory is increasingly used to improve LLM agents on recurring workplace tasks, yet its ability to produce reusable skills remains poorly understood. We introduce \benchname{}, a benchmark of 382 realistic enterprise tasks spanning six professional roles and 22 procedural skills, designed to evaluate how skills transfer across tasks, roles, and model backbones. The benchmark includes controlled evaluation settings for local improvement, cross-task transfer, cross-role transfer, and cross-model generalization.
Experiments show that procedural memory delivers consistent gains in industrial workflows: a single refinement round improves aggregate performance by 3.7-6.7 points, while skills evolved from diverse multi-model execution traces achieve 73.1\% cross-model test accuracy, outperforming all single-model trace sources. We further find that some skills generalize broadly across tasks and models, whereas others become specialized to role-specific workflows and lose effectiveness under transfer. These results provide practical guidance for building, evaluating, and deploying procedural memory systems in production agent platforms.

\end{abstract}

\section{Introduction}
\label{sec:introduction}

\begin{figure}[t]
\centering
\includegraphics[width=0.48\textwidth]{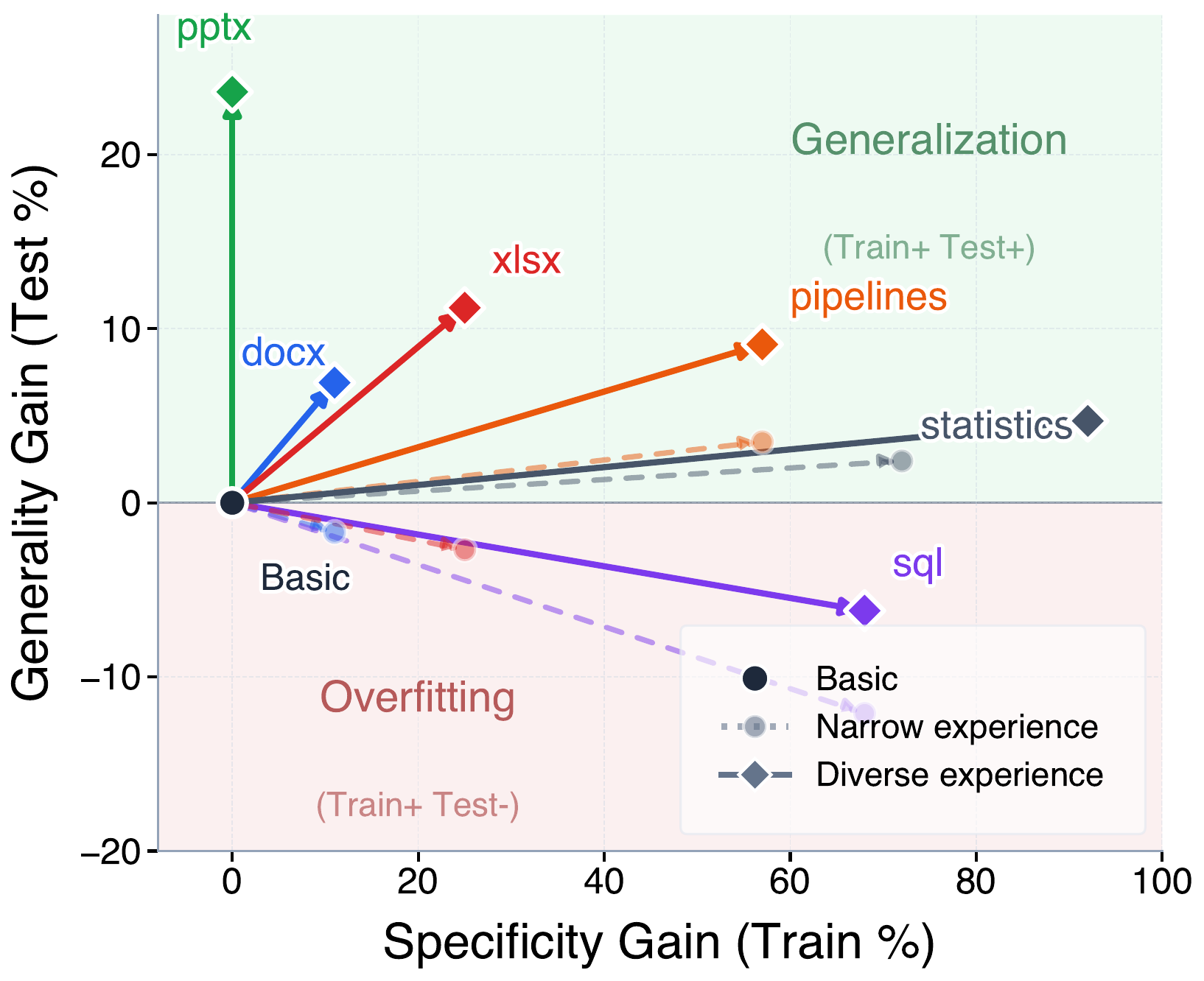}
\caption{\textbf{Skill evolution landscape.}
Procedural memories for six skills (docx, pipelines, pptx, sql, statistics, xlsx) are evolved with a Hermes memory update operator and evaluated on \benchname{}. Skills evolved from narrow experience often exhibit source-context overfitting: they improve specificity while degrading generality. Skills evolved from diverse experience move toward the desired high-specificity, high-generality regime.}
\label{fig:hero}
\end{figure}

The main AI trend of the decade is the development of LLMs~\cite{vaswani2017attention, brown2020gpt3}. Scaling training data and computation has driven broad improvements, but further scaling may face limits from bounded human-generated data~\cite{hoffmann2022chinchilla, villalobos2024run}. Meanwhile, LLM-based agents are increasingly used in practical settings~\cite{yao2023react, wang2023survey}, where they spend substantial inference-time compute on planning, tool use, reflection, and retries~\cite{reflexion, draft}. In industrial workflows, many tasks are recurring procedures rather than isolated queries: processing documents, editing spreadsheets and presentations, querying databases, configuring infrastructure, and writing tests. This creates two competing demands: cheaper and faster frameworks~\cite{skillreducer} for frequent queries in personal and corporate settings, and agents that better interact with humans and environments~\cite{expel, qian2024chatdev}, personalize to context~\cite{autoskill}, and generalize to growing task complexity~\cite{swebench, gaia}.

This shift motivates persistent mechanisms that improve reuse, reliability, and efficiency at inference time~\cite{zhang2024sprig, ramnath2025prompt}. Procedural memory is a promising direction~\cite{memp2025, mi2026procmem, wu2025agentskills}: a reusable layer of instructions, procedures, and strategies distilled from prior trajectories. For workplace agents, such memory is valuable only if it captures what transfers across tasks, users, roles, and model backbones while discarding incidental source-context details. This is difficult because trajectories depend on the model, tools, task family, and workflow that produced them; skills extracted from narrow experience may work in their source setting yet fail when the context changes~\cite{chung2024flan, kung2023active, chatterjee2025wit}.

In this work, we study procedural memory through two properties: \emph{specialization} and \emph{generalization}. Skills may be optimized for a particular workflow or learned from diverse experience spanning tasks, roles, and models. In industrial settings, the key question is whether procedural knowledge transfers beyond its source context. Diverse experience is expected to promote reusable skills, while narrow experience risks over-specialization. Figure~1 illustrates this trade-off.

Existing systems and benchmarks conflate local improvement with true transfer. Memory-augmented agent frameworks usually curate and evolve memory within a single environment, leaving generalization beyond the source setup unclear~\cite{reflexion, expel}. Skill benchmarks evaluate fixed, expert-curated skills on fixed task sets~\cite{skillsbench, skillusage}, treating skills as static artifacts rather than structures evolved from experience. General agent benchmarks contain realistic tasks, but usually lack the role--skill structure needed to test whether procedural knowledge learned in one workplace context transfers to another. The field therefore lacks a controlled setting for a practical question: \textit{when does experience produce reusable procedural structure, and when does it merely overfit to where it was observed?}

To address this, we propose:
\begin{itemize}
\item \textbf{\benchname, a benchmark for procedural skill transfer in LLM agents.}
\benchname{}\footnote{\url{https://huggingface.co/datasets/DavydenkoGr/AFTER}} contains 382 realistic workplace tasks across six professional roles and 22 procedural skills, mixing single- and multi-skill workflows, and controlled splits that measure specificity (in-context gain) and generality (held-out task, cross-role, and cross-model transfer). This structure reflects industrial agent deployments, where procedural knowledge must be reusable across recurring tasks, organizational roles, and changing model backbones.

 
\item \textbf{Empirical evidence for procedural memory transfer.} Using a transfer-oriented evaluation protocol implemented in \framework{}, we measure skill value both in the source context and under task, role, and model shifts. On the static benchmark, procedural skills improve full-pass accuracy by +2.8 points on average, while a single refinement round adds a further +5.2 points across model scales. For cross-model transfer, skills evolved from diverse multi-model traces achieve 73.1\% test accuracy, exceeding the best single-model trace source by +13.7 points. Cross-role experiments reveal a complementary limitation: skills can over-specialize to role-specific workflows and lose effectiveness when transferred across roles.
\end{itemize}

Together, these contributions recast procedural memory from a static prompt artifact into an evolving layer of agent capability that can be learned, controlled, and evaluated under realistic workplace conditions.

\section{\benchname{}: A Benchmark for Skill Transfer}
\label{sec:benchmark}

\begin{table}[h]
	\centering
	\setlength{\tabcolsep}{3pt}
	\renewcommand{\arraystretch}{1.15}
	\resizebox{\columnwidth}{!}{%
		\begin{tabular}{@{}lccccc@{}}
			\toprule
			\textbf{Benchmark}
			& \textbf{Tasks}
			& \textbf{Roles}
			& \textbf{Skills}
			& \makecell{\textbf{Multi-step}\\\textbf{tasks}}
			& \makecell{\textbf{Transfer}\\\textbf{splits}} \\
			\midrule
			GAIA & 466 & -- & \textcolor{red!60!black}{\xmark} & \textcolor{green!40!black}{\cmark} & \textcolor{red!60!black}{\xmark} \\
			SWE-bench & 2294 & 1 & \textcolor{red!60!black}{\xmark} & \textcolor{orange!60!black}{$\sim$} & \textcolor{orange!60!black}{$\sim$} \\
			SkillsBench & 85 & -- & \textcolor{green!40!black}{\cmark} & \textcolor{orange!60!black}{$\sim$} & \textcolor{red!60!black}{\xmark} \\
			WebArena & 812 & -- & \textcolor{red!60!black}{\xmark} & \textcolor{orange!60!black}{$\sim$} & \textcolor{red!60!black}{\xmark} \\
			MLE-bench & 75 & 1 & \textcolor{red!60!black}{\xmark} & \textcolor{green!40!black}{\cmark} & \textcolor{red!60!black}{\xmark} \\
			\midrule
			\textbf{\benchname{}} & 382 & 6 & \textbf{\textcolor{green!40!black}{\cmark}} & \textbf{\textcolor{green!40!black}{\cmark}} & \textbf{\textcolor{green!40!black}{\cmark}} \\
			\bottomrule
		\end{tabular}%
	}
	\vspace{2pt}
	\caption{Comparison with existing agent and skill benchmarks. \textcolor{green!40!black}{\cmark} supported, \textcolor{orange!60!black}{$\sim$} partial, \textcolor{red!60!black}{\xmark} not supported. \benchname{} uniquely combines role structure, skill annotations, and transfer splits.}
	\label{tab:comparison}
\end{table}

\begin{figure*}[h]
	\centering
	\includegraphics[width=\textwidth]{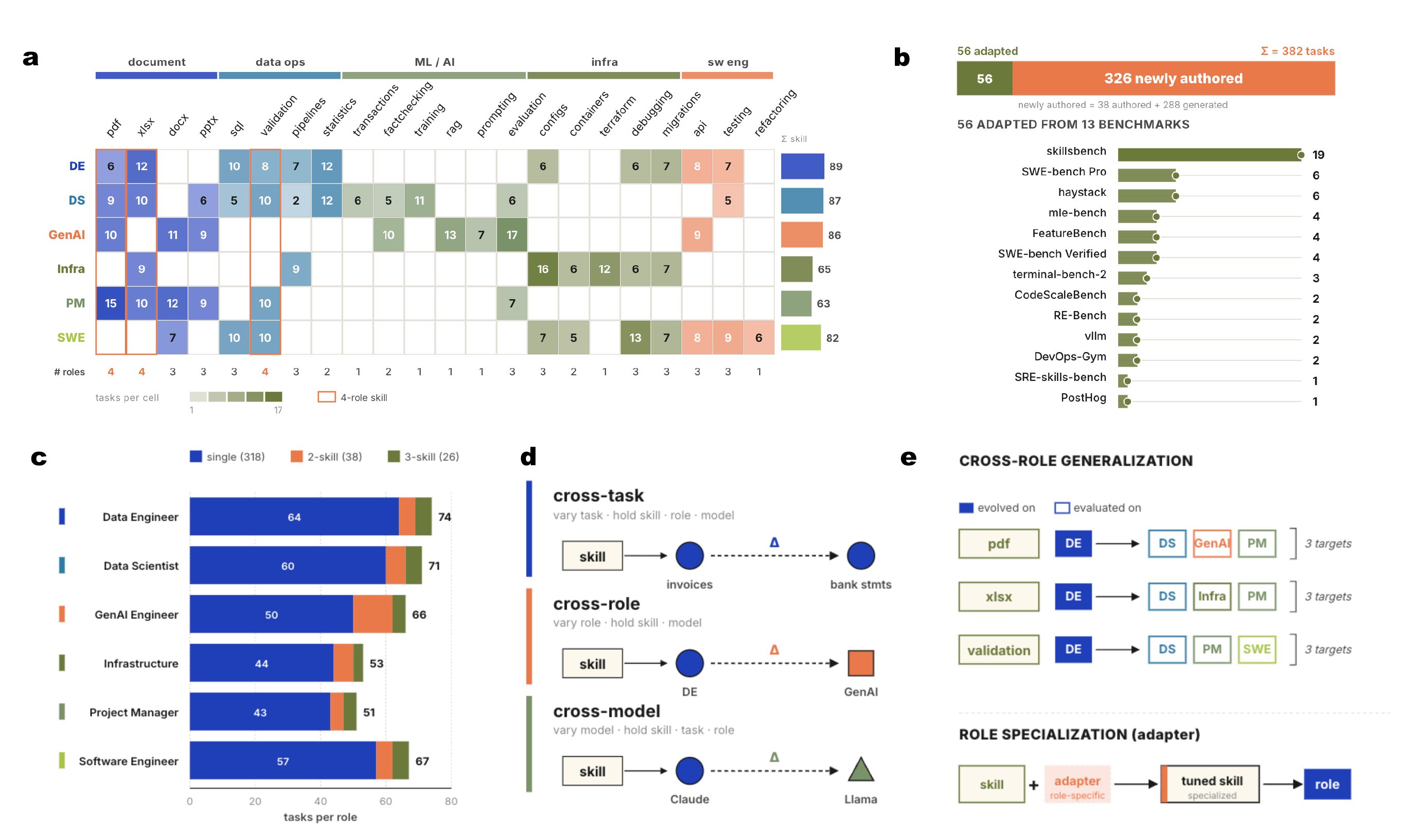}
\caption{\textbf{\benchname{} overview.}
\textbf{(a)} Role--skill matrix spanning six professional roles and five capability areas; red borders indicate skills shared across four roles.
\textbf{(b)} Task sources: 56 adapted and 326 newly authored tasks.
\textbf{(c)} Distribution of single- and multi-skill tasks by role.
\textbf{(d)} Transfer evaluation across tasks, roles, and models.
\textbf{(e)} Cross-role transfer and role-specific skill specialization.}
	\label{fig:bench}
\end{figure*}


\benchname{} is a benchmark for evaluating procedural skill transfer in LLM agents. It contains 382 realistic workplace tasks spanning six professional roles and 22 procedural skills. Unlike prior agent benchmarks that focus on task completion, \benchname{} evaluates procedural knowledge transferability across contexts. It combines a role-driven task--skill structure with controlled splits for cross-task, cross-role, and cross-model transfer (Figure~\ref{fig:bench}). Table~\ref{tab:comparison} summarizes the key differences from prior benchmarks; a detailed discussion is provided in Appendix~\ref{app:related}.

\subsection{Benchmark Design}

\paragraph{Roles.} The six roles cover common functions in technology organizations: Data Engineers (DE; data pipelines), Data Scientists (DS; statistical and ML analysis), Generative AI Engineers (GenAI; LLM applications), Infrastructure Engineers (Infra; cloud and deployment), Project Managers (PM; business documents), and Software Engineers (SWE; application code). Roles define how skills are instantiated: for example, a PDF skill may support invoice extraction for DE, document ingestion for GenAI, or executive summarization for PM. Thus, each role induces a characteristic task--skill distribution (Figure~\ref{fig:bench}a). 



\paragraph{Tasks.} Tasks split between single-skill (318) and multi-skill (64) workflows; multi-skill tasks combine two or three skills in an input--process--output structure (Figure~\ref{fig:bench}c). Skill annotations are \emph{fixed} at task definition rather than retrieved at solve time, separating skill quality from retrieval quality and giving evolution a clean optimization signal; retrieval can be studied as a separate problem on the same tasks. Data splits and the per-task file format are in Appendix~\ref{app:benchmark-details}.

\paragraph{Skills.} The catalogue covers five capability areas: document processing (pdf, xlsx, docx, pptx), data operations (sql, validation, pipelines, statistics), ML and AI (transactions, factchecking, model training, rag, prompting, evaluation), infrastructure (configs, containers, Terraform, debugging, migrations), and software engineering (api, testing, refactoring). Each skill is a self-contained \texttt{SKILL.md} artifact in the Agent Skills format~\citep{wu2025agentskills}, a versionable, retrievable unit of procedural memory at uniform granularity.

\subsection{Benchmark Construction}

Tasks come from two high-level sources. \emph{Adapted tasks} are drawn from SkillsBench~\cite{skillsbench}, SWE-bench Verified~\cite{swebench} and Pro~\cite{swebenchpro}, MLE-bench~\cite{mlebench}, FeatureBench~\cite{featurebench}, RE-Bench~\cite{rebench}, Terminal-Bench~\cite{terminalbench}, CodeScaleBench\cite{codescalebench}, DevOps-Gym~\cite{devopsgym}, SRE-skills-bench\cite{sreskillsbench}, and issues from popular open-source repositories. We preserve the core problem and success criterion, but rewrite each instruction as a self-contained workplace request and re-implement verification as a pytest suite. \emph{Newly designed tasks} cover scenarios not available in prior benchmarks; they are either practitioner-designed or first drafted with a frontier LLM and then expert-refined into realistic workplace workflows, including longer tasks requiring multiple reasoning and tool-use steps. All tasks pass automated validation and independent expert review for verifier robustness, clarity, skill fit, realism, and oracle leakage. Appendix~\ref{app:construction} details task origins, adaptation, and quality control.

In parallel, we curate 22 reusable skills from common workplace procedures in document processing, data operations, ML and AI, infrastructure, and software engineering. Each task is assigned the minimal skill set required for completion, keeping task--skill annotations fixed and separating skill quality from retrieval quality. Each skill has two prompt bodies: a handcrafted baseline (H) adapted from public skill sources and an LLM-generated body (G) drafted as a broader procedural reference (Appendix~\ref{app:examples}). This enables a controlled comparison between expert-derived and automatically authored procedural knowledge.

\subsection{Evaluation Protocol}

We evaluate skills through \emph{specificity} (source-context improvement) and \emph{generality} (transfer under distribution shift) and report two accuracy metrics. Let $\text{passed}*{k,t}$ denote the number of tests passed on attempt $k$ of task $t$, and $\text{total}*{t}$ the total number of tests.

\begin{equation}\label{eq:m1}
\text{M1}=\frac{1}{N_{\text{tasks}}}\sum_t \frac{1}{N_{\text{att}}}\sum_k
\frac{\text{passed}_{k,t}}{\text{total}_t}
\end{equation}

\begin{equation}\label{eq:m2}
\text{M2}=\frac{1}{N_{\text{tasks}}}\sum_t \frac{1}{N_{\text{att}}}\sum_k
\mathbb{1}[\text{passed}_{k,t}=\text{total}_t]
\end{equation}

M1 measures partial progress, while M2 measures complete task success.

\section{Methods}
\label{sec:methods}
\begin{table*}[h!]
	\centering
	\scriptsize
	\setlength{\tabcolsep}{3.5pt}
	\begin{tabular}{@{}ll|ccc|c|ccc|c|ccc|c|ccc|c|ccc@{}}
		\toprule
		& & \multicolumn{4}{c|}{\textbf{DS}} & \multicolumn{4}{c|}{\textbf{GenAI}} & \multicolumn{4}{c|}{\textbf{PM}} & \multicolumn{4}{c|}{\textbf{Infra}} & \multicolumn{3}{c}{\textbf{Aggregate}} \\
		\textbf{Model} & \textbf{Size} & $\varnothing$ & H & G & $\Delta$ & $\varnothing$ & H & G & $\Delta$ & $\varnothing$ & H & G & $\Delta$ & $\varnothing$ & H & G & $\Delta$ & $\varnothing$ & Best & $\Delta$ \\
		\midrule
		GPT 5.4 & L & 42.0 & 47.0 & \textbf{55.0} & \cellcolor{highgain}+13.0 & 40.0 & \textbf{43.1} & 40.0 & \cellcolor{medgain}+3.1 & 38.5 & \textbf{44.6} & 33.8 & \cellcolor{highgain}+6.1 & 50.0 & 48.7 & \textbf{56.3} & \cellcolor{highgain}+6.3 & 47.6 & 50.1 & \cellcolor{lowgain}+2.5 \\
		GPT 5.4 Mini 4 & M & 43.0 & 33.0 & \textbf{48.0} & \cellcolor{highgain}+5.0 & 35.8 & 31.6 & \textbf{45.3} & \cellcolor{highgain}+9.5 & 27.7 & 24.6 & \textbf{30.8} & \cellcolor{medgain}+3.1 & \textbf{60.0} & 43.7 & 48.7 & \cellcolor{reggain}-11.3 & 44.0 & 44.9 & \cellcolor{lowgain}+0.9 \\
		DeepSeek V4 Flash & L & 28.7 & 29.6 & \textbf{33.9} & \cellcolor{highgain}+5.2 & 31.6 & 30.5 & \textbf{35.8} & \cellcolor{medgain}+4.2 & 11.4 & 17.1 & \textbf{18.1} & \cellcolor{highgain}+6.7 & 48.8 & 46.3 & 41.2 & \cellcolor{reggain}-7.6 & 34.6 & 37.1 & \cellcolor{lowgain}+2.5 \\
		Nemotron 3 120B & M & 29.6 & \textbf{31.3} & 28.7 & \cellcolor{lowgain}+1.7 & 42.1 & 36.8 & 35.8 & \cellcolor{reggain}-6.3 & 14.3 & 21.9 & \textbf{24.8} & \cellcolor{best}+10.5 & 37.5 & 31.2 & \textbf{41.3} & \cellcolor{medgain}+3.8 & 31.9 & 33.0 & \cellcolor{lowgain}+1.1 \\
		\midrule
		Gemma 4 31B & M & 45.5 & 45.0 & \textbf{49.5} & \cellcolor{medgain}+4.0 & 35.3 & 35.3 & 33.7 & \cellcolor{reggain}-1.6 & 10.0 & 13.9 & \textbf{23.9} & \cellcolor{best}+13.9 & 43.1 & \textbf{45.6} & 44.4 & \cellcolor{lowgain}+2.5 & 38.5 & 41.3 & \cellcolor{lowgain}+2.8 \\
		Gemma 4 26B A4B & M & 37.5 & 42.0 & \textbf{48.0} & \cellcolor{best}+10.5 & 37.4 & \textbf{40.0} & \textbf{40.0} & \cellcolor{medgain}+2.6 & 16.9 & \textbf{24.6} & 20.0 & \cellcolor{highgain}+7.7 & 34.4 & 28.1 & \textbf{40.0} & \cellcolor{highgain}+5.6 & 36.2 & 39.7 & \cellcolor{medgain}+3.5 \\
		Gemma 4 E4B & S & 20.5 & 18.5 & \textbf{24.5} & \cellcolor{medgain}+4.0 & 15.3 & 23.7 & \textbf{29.5} & \cellcolor{best}+14.2 & 8.5 & 11.5 & \textbf{13.1} & \cellcolor{highgain}+4.6 & 13.8 & 14.4 & \textbf{21.2} & \cellcolor{highgain}+7.4 & 19.4 & 24.0 & \cellcolor{highgain}+4.6 \\
		\midrule
		Qwen 3.5-397B-FP8 & L & 36.0 & 35.2 & \textbf{40.2} & \cellcolor{medgain}+4.2 & 35.3 & 35.5 & \textbf{38.7} & \cellcolor{medgain}+3.4 & 12.3 & 13.5 & \textbf{16.9} & \cellcolor{highgain}+4.6 & 48.1 & 46.9 & 45.0 & \cellcolor{reggain}-3.1 & 37.8 & 41.3 & \cellcolor{medgain}+3.5 \\
		Qwen 3.5-122B-A10B & L & 36.0 & 33.5 & \textbf{41.0} & \cellcolor{highgain}+5.0 & 30.0 & 34.2 & \textbf{35.3} & \cellcolor{highgain}+5.3 & 12.3 & 19.2 & \textbf{20.8} & \cellcolor{highgain}+8.5 & 44.4 & \textbf{45.0} & 42.5 & \cellcolor{lowgain}+0.6 & 36.5 & 39.7 & \cellcolor{medgain}+3.2 \\
		Qwen 3.5-35B-A3B & M & 21.5 & 25.5 & \textbf{30.5} & \cellcolor{highgain}+9.0 & 27.9 & 31.6 & \textbf{36.9} & \cellcolor{highgain}+9.0 & 13.1 & 13.1 & \textbf{14.6} & \cellcolor{lowgain}+1.5 & 28.1 & 23.7 & \textbf{35.0} & \cellcolor{highgain}+6.9 & 26.9 & 32.2 & \cellcolor{highgain}+5.3 \\
		Qwen 3.5-9B & S & 12.5 & 11.5 & \textbf{17.0} & \cellcolor{highgain}+4.5 & 17.9 & 15.3 & \textbf{18.9} & \cellcolor{lowgain}+1.0 & 6.2 & 10.8 & \textbf{14.6} & \cellcolor{highgain}+8.4 & 11.3 & 13.8 & \textbf{20.6} & \cellcolor{highgain}+9.3 & 15.7 & 18.7 & \cellcolor{medgain}+3.0 \\
		\midrule
		GPT-oss-120B & L & 44.5 & 47.0 & \textbf{49.0} & \cellcolor{highgain}+4.5 & 38.3 & \textbf{42.6} & 41.6 & \cellcolor{medgain}+4.3 & 32.3 & 30.8 & \textbf{33.1} & \cellcolor{lowgain}+0.8 & 57.5 & 51.3 & \textbf{58.7} & \cellcolor{lowgain}+1.2 & 43.2 & 45.6 & \cellcolor{lowgain}+2.4 \\
		GPT-oss-20B & M & 30.0 & \textbf{32.0} & 29.0 & \cellcolor{lowgain}+2.0 & 32.6 & 32.6 & 32.6 & \cellcolor{lowgain}+0.0 & 16.1 & \textbf{22.3} & 16.9 & \cellcolor{highgain}+6.2 & 30.6 & 25.6 & 30.0 & \cellcolor{reggain}-0.6 & 30.9 & 31.3 & \cellcolor{lowgain}+0.4 \\
		\bottomrule
	\end{tabular}
\caption{Static M2 (\%) on \benchname{} under no-skill ($\varnothing$), handcrafted (H), and generated (G) skills. $\Delta$ denotes the best gain over no-skill. Colors indicate: \colorbox{best}{$\geq$+10}, \colorbox{highgain}{+4..+10}, \colorbox{medgain}{+2..+4}, \colorbox{lowgain}{0..+2}, \colorbox{reggain}{$<$0}.}
	\label{tab:static-main}
\end{table*}

\subsection{Procedural-Memory Optimization}
\label{sec:formal-objective}

Let $\Sigma \subset \mathcal{S}$ be a procedural-memory configuration (a single skill or a
skill library) and $\mathcal{D}=\{\tau_i\}_{i=1}^{N}$ a pool of $N$ traces collected under
$\Sigma$ or its earlier versions. An update operator $U$ maps experience to a new
configuration, $\Sigma' = U(\Sigma,\mathcal{D})$, and may be instantiated as reflection,
distillation, or a learned memory-writing policy~\citep{seasurvey}. Given an initial
$\Sigma_0$, source and target distributions $p_{\mathrm{src}},p_{\mathrm{tgt}}$, and a trace
budget $N$, we seek the update rule maximizing expected value on target contexts:
\begin{equation*}
\label{eq:procmem-opt}
\begin{aligned}
U^{*} = \arg\max_{U \in \mathcal{U}}\;
\mathbb{E}_{\substack{\mathcal{D}_{\mathrm{src}} \sim p_{\mathrm{src}} \\ c \sim p_{\mathrm{tgt}}}}
&\bigl[\, V\bigl( U(\Sigma_0, \mathcal{D}_{\mathrm{src}});\, c \bigr) \,\bigr] \\
\text{s.t.}\quad &|\mathcal{D}_{\mathrm{src}}| \leq N .
\end{aligned}
\end{equation*}
Here $\mathcal{U}$ is the admissible family of update mechanisms and $V(\Sigma;c)$ is the
value of configuration $\Sigma$ in context $c$. The protocol sets
$p_{\mathrm{tgt}}=p_{\mathrm{src}}$ for \emph{specificity} and shifts the task, role, or model
distribution for \emph{generality}.

\subsection{\framework{}: Benchmark Evaluation Interface}
\label{sec:framework}
To compare procedural-memory systems on \benchname{}, we use \framework{}, a lightweight
harness that standardizes trace collection, skill versioning, update execution, and transfer
measurement. Skills are stored as versioned \texttt{SKILL.md} artifacts with YAML metadata and
markdown bodies; each execution emits a trace linked to the active skill version, making updates
and evaluations reproducible.
\framework{} supports full-skill updates through a
\textsc{Collect}--\textsc{Diagnose}--\textsc{Revise}--\textsc{Promote} cycle. We denote by
$\rho$ the \emph{reflector}: the model or procedure that inspects traces, summarizes failure
modes, and proposes a revised skill body. For current skill version $s^{(v)}$ and source trace
pool $\mathcal{D}_{\mathrm{src}}$, the update is
$s^{(v+1)} = U_{\rho}(s^{(v)}, \mathcal{D}_{\mathrm{src}})$.
Thus, \framework{} fixes trace collection, validation, promotion, rollback, and lineage
tracking, while the reflector $\rho$ may vary across experiments or external systems. We
evaluate four external procedural-memory systems through this interface; their update mechanisms
are summarized in Appendix~\ref{app:external-systems}. Full harness details are in
Appendix~\ref{app:operators}.

\section{Experiments and Results}
\label{sec:experiments}

We evaluate procedural memory on \benchname{} in four stages. First, we measure the value of static skill content without adaptation. Second, we test whether a single refinement pass can improve existing skills. Third, we compare trace-based skill evolution under narrow and diverse experience. Finally, we analyze transfer across models and roles, together with inference efficiency. Full experimental details are provided in Appendix~\ref{app:experimental-setup}.

\subsection{Static Skill Valuation}

In the static setting, each LLM is invoked once per task with the task instruction and, optionally, a skill in the prompt; there is no agent orchestration, retrying, tool use, or evolution. Skill content takes one of three forms: none ($\varnothing$), handcrafted (H), or LLM-generated (G). We report M2 as the primary metric, M1 results are provided in Appendix Table~\ref{tab:pass-rate-train-subset}.

Table~\ref{tab:static-main} reports results for the four roles most affected by skill availability (DS, GenAI, PM, Infra), along with aggregate statistics. Full per-role results are provided in Appendix Table~\ref{tab:perfect-rate-train-subset}. Skills benefit weaker models more consistently than frontier models: for example, Gemma 4 E4B gains +14.2 points on GenAI, while GPT 5.4 gains +3.1. LLM-generated skills (G) often outperform handcrafted skills (H), suggesting that automated skill authoring can match expert-derived procedural knowledge. Gains vary by role: DS and GenAI benefit most, whereas DE and SWE show smaller improvements, likely because coding-heavy roles already perform well without explicit procedural guidance.

\subsection{LLM-Guided Skill Improvement}
\begin{figure}[h!]
	\centering
	\includegraphics[width=\columnwidth]{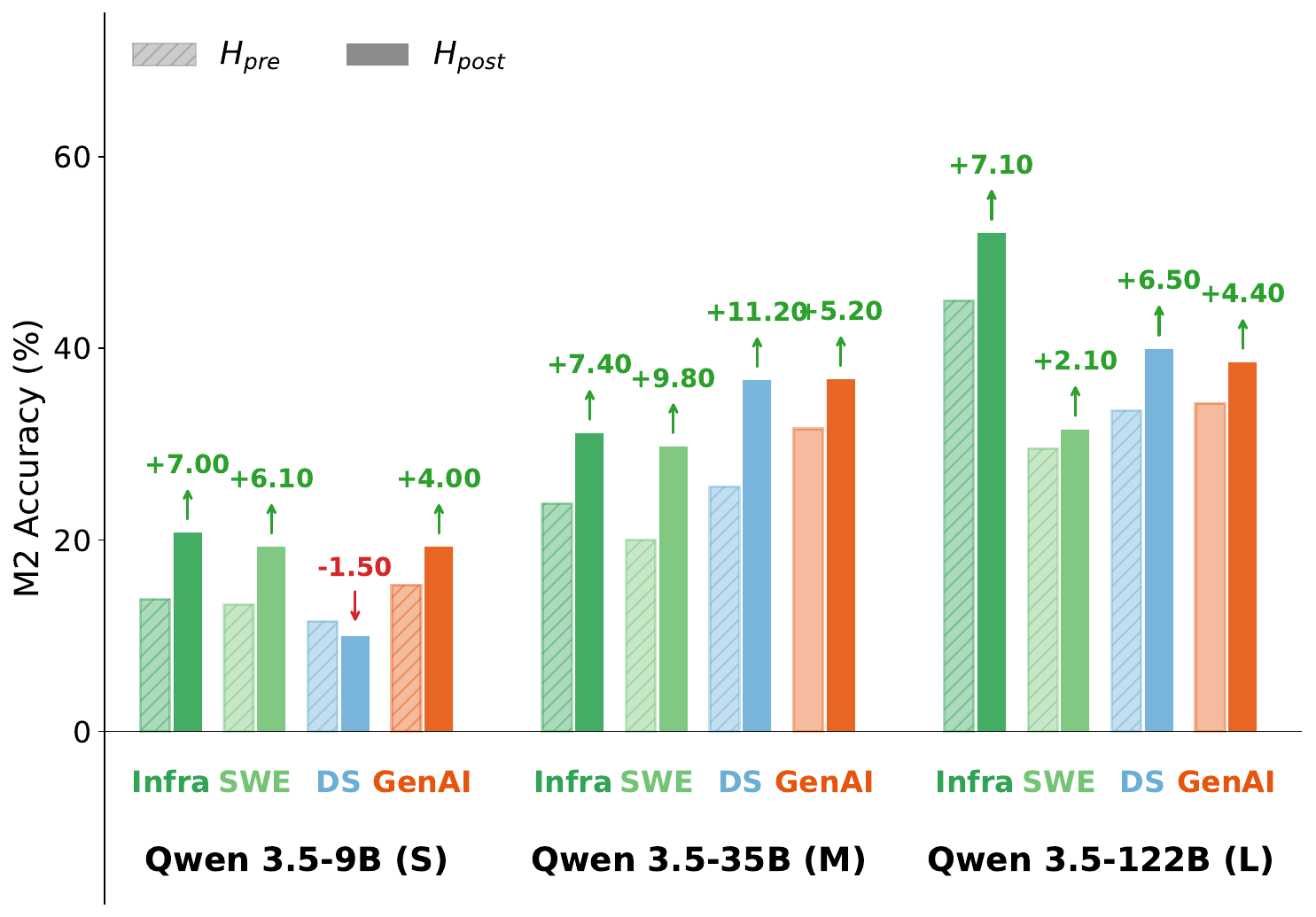}
	\caption{Single-round refinement impact: M2 accuracy before ($H_{pre}$) and after ($H_{post}$) for top-4 roles by gain.}
	\label{fig:refine-h}
\end{figure}

The static setup isolates the value of skill \emph{content} but says nothing about how procedural memory should be \emph{produced} or \emph{adapted}. Before exploring multi-round evolution, we test whether a single refinement pass can improve existing skills. We apply one round of LLM-guided refinement (using Evolution with Codex as reflector) to the handcrafted skill catalogue, producing H$_{\text{post}}$ from H$_{\text{pre}}$.

Figure~\ref{fig:refine-h} shows that even a single refinement round yields consistent improvements: +3.7 to +6.7 aggregate points across model scales. Larger models benefit more consistently, with Infra and SWE showing the strongest gains. Full results are in Appendix Table~\ref{tab:e1-refine-h}.

\subsection{Framework-Guided Skill Improvement}

\begin{table}[H]
\centering
\footnotesize
\setlength{\tabcolsep}{4pt}
\renewcommand{\arraystretch}{1.15}
\begin{tabular}{lrrrrr}
\toprule
\textbf{Framework} &
\textbf{Seed} &
\multicolumn{2}{c}{\textbf{Narrow}} &
\multicolumn{2}{c}{\textbf{Diverse}} \\
\cmidrule(lr){3-4}\cmidrule(lr){5-6}
& &
$\Delta_{\rm tr}$ & $\Delta_{\rm te}$ &
$\Delta_{\rm tr}$ & $\Delta_{\rm te}$ \\
\midrule
Codex GPT-5.5 & 57.1 & -1.7 & \cellcolor{lowgain}+0.4 & -2.5 & \cellcolor{highgain}+8.3 \\
Hermes & 58.4 & +3.6 & \cellcolor{reggain}-1.4 & +3.7 & \cellcolor{best}+18.0 \\
Memento & 52.4 & -12.7 & \cellcolor{lowgain}+0.1 & +3.8 & \cellcolor{lowgain}+2.0 \\
MemP & 56.6 & +13.9 & \cellcolor{medgain}+2.5 & +7.3 & \cellcolor{reggain}+0.0 \\
EvoSkill & 52.5 & +14.9 & \cellcolor{reggain}-2.7 & -5.7 & \cellcolor{reggain}-3.9 \\
\bottomrule
\end{tabular}
  \caption{Train and test M1 gains over handcrafted skills on pdf, xlsx, and pptx tasks using a shared Qwen3.5-35B-A3B solver, averaged over pdf, xlsx, and pptx.
           Seed = handcraft test M1; $\Delta_\text{tr}$ / $\Delta_\text{te}$ = evolved$-$seed on train / test (same task set per condition).
           Trace and skill management are framework-specific. Narrow: $n=1$; Diverse: $n=5$. Colors denote test gains.}
  \label{tab:evolution_gains}
\end{table}

We evaluate procedural memory frameworks on skill evolution from execution traces. Table~\ref{tab:evolution_gains} compares five approaches under narrow ($n=1$) and diverse ($n=5$) evolution. We run frameworks through the same Evolution harness under identical conditions with shared Qwen3.5-35B-A3B solver, task pool, and train/test splits. Results reveal a clear gap between specialization and transfer: all frameworks struggle with the proper generalization, large training gains do not necessarily translate into improvements on held-out tasks.

\subsection{Skill Transfer Analysis}

We analyze skill transfer along three practically important dimensions: cross-model generalization, cross-role transfer, and inference efficiency.

\begin{figure}[h]
	\centering
	\includegraphics[width=\columnwidth]{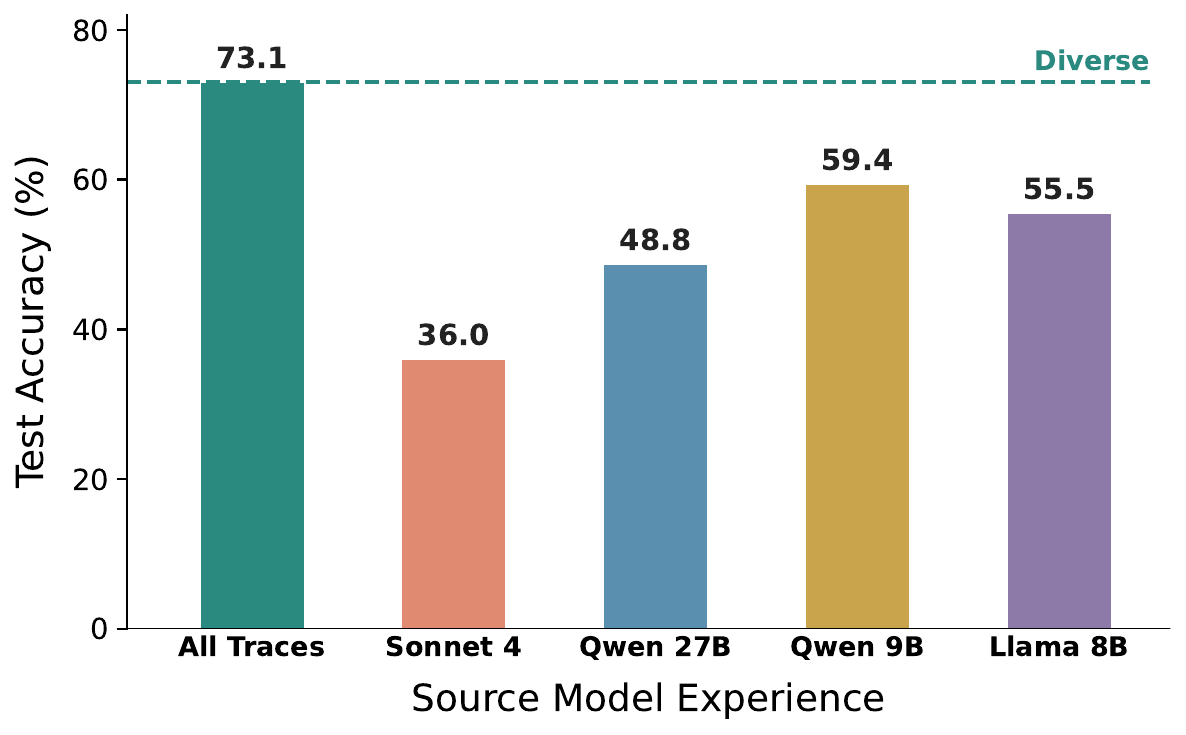}
	\caption{Cross-model transfer: test accuracy when skills are evolved from traces of different source models versus diverse traces from all models.}
	\label{fig:model-experience}
\end{figure}

\paragraph{Cross-model transfer.} A key question is whether skills must be evolved from traces of the same model that will use them. Figure~\ref{fig:model-experience} shows that skills evolved combining experience from multiple source models substantially outperform single-model sources: 73.1\% versus 36.0--59.4\%. Surprisingly, weaker source models provide better transferable signal than stronger models, suggesting that procedural knowledge benefits from imperfect executions.

\begin{figure}[h]
	\centering
	\includegraphics[width=1\columnwidth]{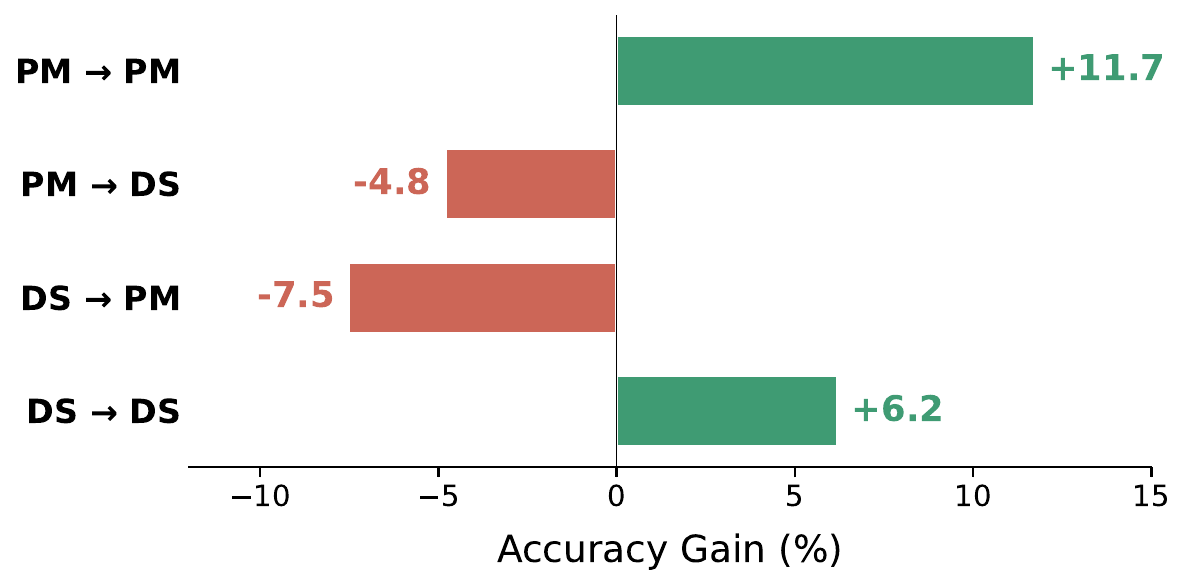}
	\caption{Cross-role transfer for the \texttt{pdf} skill. In-role evolution (PM to PM, DS to DS) yields gains, while applying a skill evolved for one role to another (PM to DS, DS to PM) hurts performance.}
	\label{fig:cross-role}
\end{figure}

\paragraph{Cross-role generalization.} Skills evolved within one professional role may not transfer effectively to other roles. Figure~\ref{fig:cross-role} illustrates this for the pdf skill: while in-role evolution yields gains of +11.7 (PM) and +6.2 (DS), applying a skill evolved for one role to another produces losses of -4.8 to -7.5 points. This asymmetry arises because different roles use the same skill for different purposes (e.g., pdf extraction for executive summaries in PM versus data ingestion in DS). Role-specific specialization emerges naturally during evolution.

\begin{figure}[h]
	\centering
	\includegraphics[width=\columnwidth]{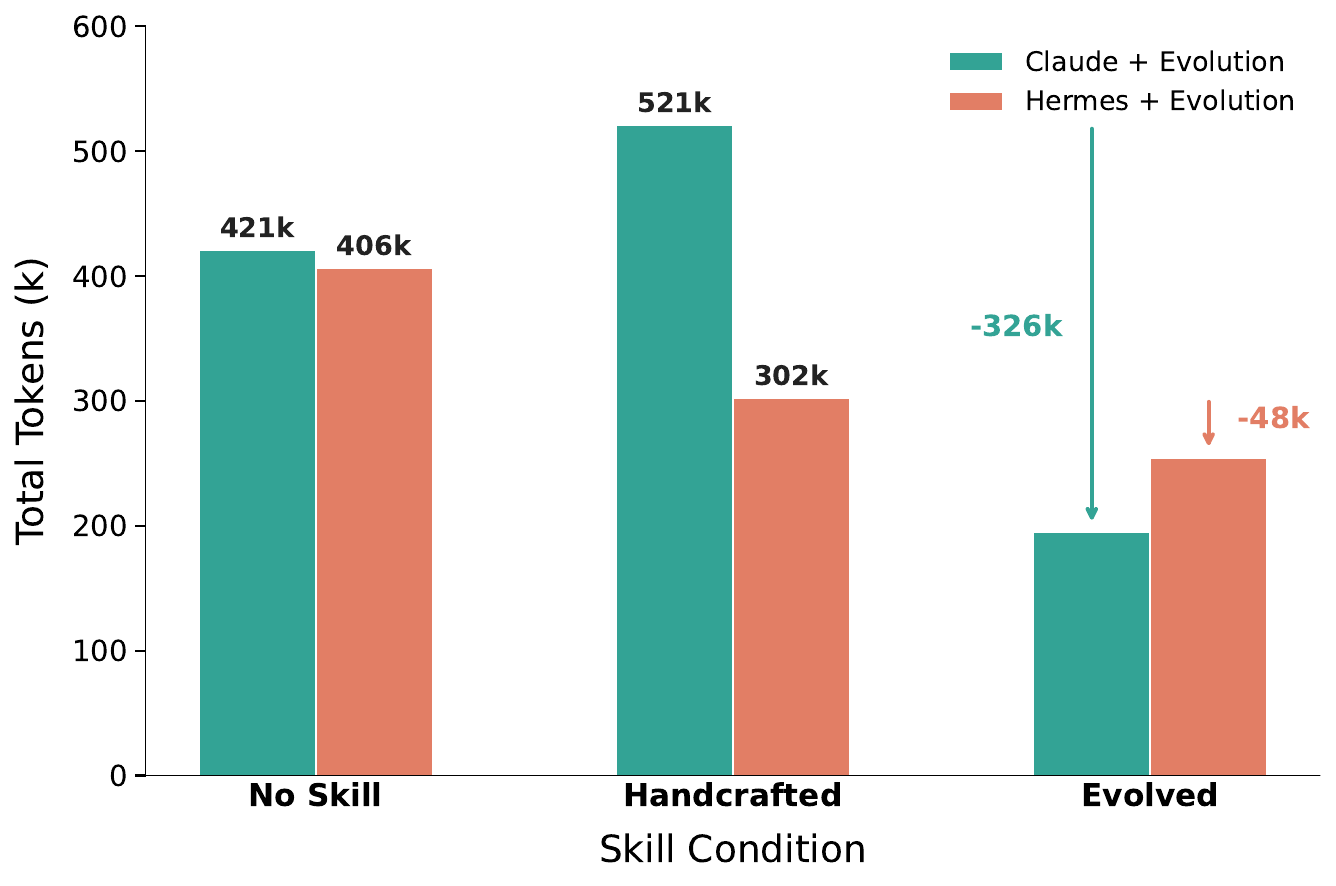}
	\caption{Token usage for Kafka Lag Anomaly Detection. Evolved skills reduce total tokens by 326k (Claude) and 48k (Hermes) compared to handcrafted skills.}
	\label{fig:token-usage}
\end{figure}

\paragraph{Token efficiency.} Evolved skills reduce inference cost by front-loading procedural knowledge into the prompt rather than discovering it at runtime. Figure~\ref{fig:token-usage} illustrates this on the Kafka Lag Anomaly Detection task: evolved skills reduce total token usage by 326k (62\%) for Claude and 48k (16\%) for Hermes compared to handcrafted skills. The same pattern holds across other tasks, with evolved skills consistently reducing both token usage and inference cost (Table~\ref{tab:claude-hermes-usage}).



\section{Conclusion}
\label{sec:conclusion}

We introduced \benchname{}, a benchmark for evaluating procedural memory through transfer across tasks, roles, and model backbones. Across 382 workplace tasks, we find that procedural skills improve full-pass accuracy by +2.8 points on average, while a single round of skill evolution yields an additional +5.2-point gain. Skills evolved from diverse multi-model traces achieve 73.1\% cross-model test accuracy, outperforming the best single-model trace source by at least +13.7 points. At the same time, cross-role studies reveal an important limitation: skills naturally specialize to local workflows and may lose effectiveness when transferred across contexts.
Taken together, these results suggest that the central challenge of procedural memory is not storing more experience, but extracting procedural structure that remains useful beyond the environment in which it was learned.

\section*{Limitations}
\label{sec:limits}

\textbf{Benchmark coverage}. \benchname{} targets technology-sector roles and workplace tasks drawn partly from authors' practice, which may underrepresent domains such as healthcare, legal, or scientific research. The 22 skills span five capability areas but intentionally exclude open-ended creative or conversational tasks, limiting conclusions to procedural, tool-use-oriented workflows.

\textbf{Evaluation scope}. Our experiments fix the trace budget per evolution run to enable controlled comparison; real deployments may accumulate far larger trace pools, and the relationship between trace volume and transfer quality remains an open question. Evaluation uses automated pytest verification, which measures functional correctness but does not capture qualities such as code readability, robustness to edge cases beyond the test suite, or user preference.

\textbf{Model and framework selection}. We evaluate a representative but non-exhaustive set of LLMs and procedural-memory frameworks. Several recently released frontier models and memory systems could not be included due to API access constraints or release timing relative to benchmark finalization. 

\section*{Ethics Statement}

All datasets used are public and were collected and preprocessed by their original authors.

\bibliography{references}

\clearpage
\appendix
\section*{Appendix}
\appendix
\counterwithin{table}{section}

\section{Related Work}
\label{app:related}

\paragraph{Skills as reusable procedural artifacts.}
LLM agents can improve from experience without weight updates by converting interaction traces into reusable guidance. Early work explored verbal self-reflections~\citep{reflexion}, cross-task insights and replayed successes~\citep{expel}, tool documentation rewritten from trial-and-error~\citep{draft}, and directly optimised prompts~\citep{opro}. A second wave externalised that guidance into discrete, model-agnostic skill artifacts~\citep{autoskill, learnact}, augmented with explicit storage and update policies~\citep{memp2025}, MDP-style formalisation~\citep{mi2026procmem}, or hierarchical refinement~\citep{skillx}. Persistent skill files now appear in deployed coding assistants~\citep{cursorstudy} and as structured blueprints in autonomous-network agents~\citep{hermes}, marking procedural memory as a first-class artifact outside the base prompt~\citep{wu2025agentskills}.

\paragraph{Self-evolving procedural memory.}
Once skills are first-class artifacts, the natural next question is how they should change as the agent accumulates experience. Memento-Skills couples continual skill writing with behaviour-aligned routing~\citep{mementoskills}; EvoSkill and CoEvoSkills refine skills from failure and verification feedback~\citep{evoskill, coevoskills}; SkillClaw aggregates cross-user trajectories into a shared repository~\citep{skillclaw}; and broader self-evolving-agent work targets agent code or training loops~\citep{dgm, seasurvey}. These methods establish that evolution is feasible, but each fixes a single source-pool design and reports transfer along at most one axis, leaving it unclear whether the resulting behaviour generalises beyond the source setup.

\paragraph{Agent and skill benchmarks.}
Measuring such transfer, in turn, requires benchmarks that isolate procedural memory from the rest of the agent stack. End-to-end agent benchmarks --- GAIA~\citep{gaia}, SWE-bench~\citep{swebench}, WebArena~\citep{webarena}, and MLE-bench~\citep{mlebench} --- score full pipelines without separating the skill artifact from planning, retrieval, or tool use. Skill-focused benchmarks come closer: SkillsBench compares no-skill, curated-skill, and self-generated-skill conditions under controlled verifiers~\citep{skillsbench}, and follow-up work shows that retrieval from large noisy corpora collapses much of the gain~\citep{skillusage}. Yet none of these varies the skill source while holding the rest fixed, and none combines explicit skill annotations, professional-role structure, and controlled transfer splits in a single setting (Table~\ref{tab:comparison}).


\section{Evolution Details}
\label{app:operators}

\begin{figure*}[t]
	\centering
	\includegraphics[width=\textwidth]{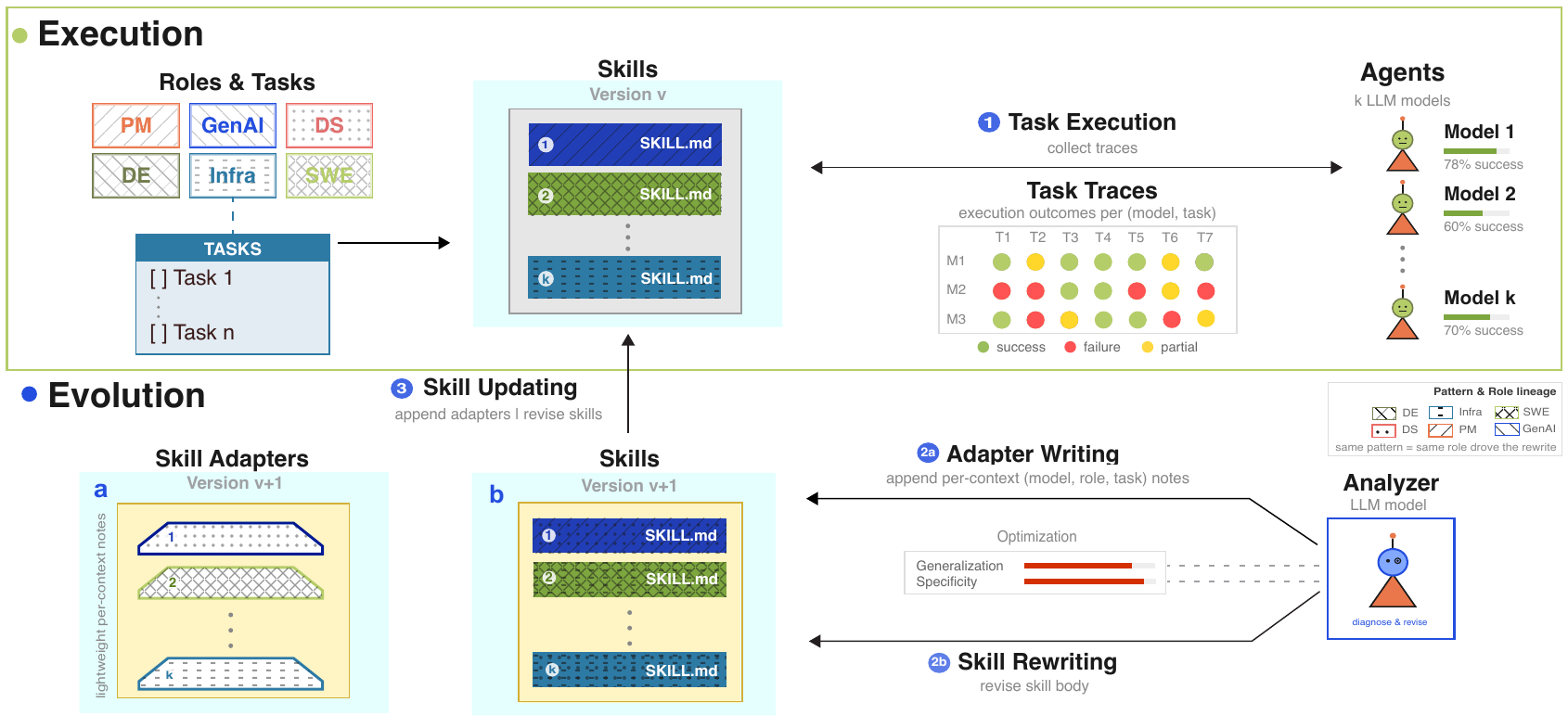}
	\caption{
	The \framework{} pipeline. Agent executions emit traces associated with an active skill version. Traces support diagnosis, revision, and validation of candidate versions. Accepted and rejected candidates remain linked in a lineage graph. Context-specific adapters can specialize a frozen skill body for a task, role, or model without modifying the main body.
	}
	\label{fig:evolution-pipeline}
\end{figure*}

\framework{} is a lightweight framework for controlling, recording, and evaluating procedural-memory evolution. It is used as an experimental layer around agentic systems with textual procedural memory. Its purpose is to make each skill version, trace, update, evaluation, promotion, and rollback explicit and reproducible, so that different update mechanisms and trace sources can be studied under the same benchmark protocol (Figure~\ref{fig:evolution-pipeline}).

\paragraph{Skill representation.}
Each skill is stored as a versioned textual artifact. In our implementation, a \texttt{SKILL.md} file contains YAML metadata and a markdown body. The metadata records the skill name, role and skill annotations, version, parent version, source trace pool, and evaluation status. The body contains the procedural content used by the agent; Appendix~\ref{app:examples} shows an example.

\paragraph{Traces and lineage.}
Each execution under version $s^{(v)}$ emits a trace $\tau$ linked to that version. Traces provide evidence for failure diagnosis, revision, and fitness estimation. When an operator modifies a skill, \framework{} creates a new version $s^{(v+1)}$ and links it to its parent. Rejected candidates remain as inactive branches. Named snapshots store sets of active skill versions, making benchmark runs reproducible.

\paragraph{Full-skill evolution.}
One iteration follows four stages: \textsc{Collect}, \textsc{Diagnose}, \textsc{Revise}, and \textsc{Promote/Rollback}. The agent first executes source tasks with the current skill. The framework then aggregates failure traces into recurrent error modes, such as missing checks, brittle assumptions, incorrect tool use, or incomplete output requirements. A revision operator proposes a candidate body conditioned on the current skill and the diagnosis. The candidate is promoted if it improves validation performance by at least margin $\delta$; otherwise it is retained as an inactive branch. For current version $s^{(v)}$ and source trace pool $\mathcal{D}_{\mathrm{src}}$, the body-level update is $s^{(v+1)}=U(s^{(v)},\mathcal{D}_{\mathrm{src}})$. The framework fixes trace collection, validation, promotion, rollback, and lineage tracking, while the revision operator itself can vary.

\paragraph{Context-specific adapters.}
An alternative update strategy keeps the skill body frozen and instead prepends a short context-specific prefix distilled from execution traces. Such adapters could be keyed by task, role, or model, separating local specialization from body-level evolution without modifying shared procedural content. We leave their empirical evaluation to future work.

\section{External Agentic Frameworks with Procedural Memory}
\label{app:external-systems}

We evaluate four procedural-memory systems through the \framework{} interface. They differ in how procedural knowledge is represented, selected, and updated.

\paragraph{EvoSkill.}
EvoSkill~\citep{evoskill} refines skills from failure and verification feedback. It uses a generate--verify--refine loop in which successful patterns are reinforced and failing patterns are corrected across iterations.

\paragraph{Memp.}
Memp~\citep{memp2025} distills trajectories into fine-grained stepwise instructions and higher-level script-like procedures. Its memory loop separates build, retrieval, and update phases, with update operations that add, modify, or delete memory entries in response to execution feedback.

\paragraph{Hermes.}
Hermes~\citep{hermes} represents procedural knowledge as structured YAML blueprints generated and refined in a multi-agent Designer/Coder pipeline. Candidate plans are critiqued, merged, and edited using evaluator judgments and execution feedback before code generation.

\paragraph{Memento-Skills.}
Memento-Skills~\citep{mementoskills} stores procedural memory as structured markdown skills embedded in stateful prompts. It combines a behaviour-trainable router for skill selection with online skill writing and library expansion, allowing skill selection and skill editing to co-evolve without parameter updates.

\section{Benchmark Details}
\label{app:benchmark-details}

\subsection{Data Splits}

Each (role, skill) cell in the benchmark has tasks split into three folds: train (50\%), validation (25\%), and test (25\%). The train split provides traces for skill evolution. The validation split enables hyperparameter tuning and prevents overfitting during evolution. The test split provides a final evaluation, ensuring reported numbers reflect genuine generalization.

For cross-role experiments, an additional split structure holds out entire roles. When measuring cross-role transfer for the PDF skill, we might train on DE-PDF, DS-PDF, and PM-PDF tasks, then evaluate on GenAI-PDF tasks that were never seen during evolution.

\subsection{Task Format}

Each task in \benchname{} follows a consistent structure:
\begin{itemize}
    \item \textbf{Metadata} (\texttt{task.toml}): task name, role, required skills, difficulty, and data source attribution.
    \item \textbf{Instructions} (\texttt{instruction.md}): realistic request mimicking how a colleague might describe the work.
    \item \textbf{Input files} (\texttt{inputs/}): authentic data in workplace formats (Excel, PDF, CSV, JSON).
    \item \textbf{Verification} (\texttt{tests/test\_outputs.py}): pytest assertions for automated correctness checking.
\end{itemize}

Task instructions deliberately omit implementation details that should come from procedural memory. For example, an instruction might say, ``extract the transaction table* from the bank statement'' without specifying which library to use, how to handle multi-page documents, or the output format. These procedural details should be supplied by the skill.

\section{Benchmark Construction}
\label{app:construction}

\subsection{Task Origins}
\label{app:origins}

\benchname{} draws tasks from three complementary sources (Table~\ref{tab:construction-origins}): tasks adapted from public benchmarks, tasks written by the authors, and tasks produced through multi-stage LLM-based generation.

\begin{table*}[h]
\centering
\small
\begin{tabular}{@{}lrl@{}}
\toprule
\textbf{Origin} & \textbf{\# Tasks} & \textbf{Role of this category in the benchmark} \\
\midrule
Adapted from existing public benchmarks    & 56  & External anchor; comparability with prior work \\
Written by the authors                     & 38  & Coverage of topics relevant to the authors' practice \\
LLM-based generation                       & 288 & Scaled benchmark coverage \\
\midrule
\textbf{Total active pool}                 & \textbf{382} & \\
\bottomrule
\end{tabular}
\caption{Origins of the \benchname{} task pool.}
\label{tab:construction-origins}
\end{table*}

\paragraph{Adapted tasks.}
\label{app:sources}
56 tasks were adapted from 13 public benchmarks and source repositories (Table~\ref{tab:construction-sources}). For each, we kept the underlying problem and success criterion, rewrote the instruction to conform to the \benchname{} task contract (a self-contained workplace request, no embedded oracle, no references to the upstream evaluator), and re-implemented verification as a pytest suite in the unified \benchname{} test harness. Tasks that could not be adapted faithfully were dropped rather than rewritten.

\begin{table*}[h]
\centering
\small
\setlength{\tabcolsep}{4pt}
\begin{tabular}{@{}llrl@{}}
\toprule
\textbf{Upstream source} & \textbf{Citation} & \textbf{\# Tasks} & \textbf{Roles covered} \\
\midrule
benchflow-ai/skillsbench                    & \citep{skillsbench}                   & 19 & DE, DS, Infra, PM \\
ScaleAI/SWE-bench\_Pro                      & \citep{swebenchpro}                   &  6 & DE, SWE \\
deepset-ai/haystack (issues)                & --                                    &  6 & GenAI, SWE \\
openai/mle-bench                            & \citep{mlebench}                      &  4 & DS, GenAI \\
princeton-nlp/SWE-bench Verified            & \citep{swebench}                      &  4 & SWE \\
LiberCoders/FeatureBench                    & \citep{featurebench}                  &  4 & SWE \\
harbor-framework/terminal-bench-2           & \citep{terminalbench}                 &  3 & DE, GenAI, Infra \\
sourcegraph/CodeScaleBench                  & --                                    &  2 & DE, GenAI \\
METR/RE-Bench                               & \citep{rebench}                       &  2 & DS, GenAI \\
vllm-project/vllm (issues)                  & \citep{vllm}                          &  2 & GenAI \\
ucsb-mlsec/DevOps-Gym                       & \citep{devopsgym}                     &  2 & SWE \\
Rootly-AI-Labs/SRE-skills-bench             & --                                    &  1 & Infra \\
PostHog/posthog (issues)                    & --                                    &  1 & SWE \\
\midrule
\textbf{Total adapted}                      &                                       & \textbf{56} & \\
\bottomrule
\end{tabular}
\caption{Upstream sources of adapted tasks. Entries marked ``--'' are open-source projects (GitHub issues or benchmarks) without an associated peer-reviewed or arXiv publication.}
\label{tab:construction-sources}
\end{table*}

\paragraph{Newly designed tasks.}
38 tasks were written by the authors on topics relevant to their own practice. 18 of these are longer multi-turn scenarios (3 per role) that probe procedural memory across several reasoning steps. The remaining 288 tasks were drafted by Claude Sonnet 4.6. Each draft was scored and iteratively rewritten until it met all criteria below.

\subsection{Quality Assurance}
\label{app:qa}

Every task in the pool was reviewed under a uniform protocol combining automated checks with two-reviewer inspection against the following criteria:

\begin{enumerate}
   \item[A.1] \textbf{Clarity} --- \texttt{instruction.md} is unambiguously interpretable on its own.
    \item[A.2] \textbf{Skill fit} --- success requires the declared skills.
    \item[A.3] \textbf{Realism} --- the scenario is plausible for the assigned role.
    \item[B.4] \textbf{Dependency hygiene} --- lightweight, easily installable dependencies only.
    \item[B.5] \textbf{Verifier soundness} --- the verifier rejects adversarial baselines (empty, constant, random).
    \item[B.6] \textbf{No oracle leakage} --- the instruction contains no ground-truth values, hard-coded answers, or hidden hints.
    \item[C.7] \textbf{Determinism} --- repeated runs against regenerated inputs yield consistent verifier outcomes.
    \item[C.8] \textbf{Self-containment} --- all materials are in the task directory or generated from a fixed seed.
\end{enumerate}

Automated checks operationalize B.4--B.5 and C.7--C.8: a static metadata and path-contract audit, a reference-solution calibration run, and an adversarial gate that substitutes empty, constant, and random outputs and requires the verifier to fail on all three. Two authors then independently reviewed every task against the full rubric. A task was accepted only when both reviewers returned \textit{accept} on every criterion. Disagreements led to rewriting and re-review. Total human-review effort was approximately 32 hours per reviewer.

\section{Experimental Setup}
\label{app:experimental-setup}

We evaluate on \benchname{} across all six professional roles and the full 22-skill catalogue, following the train/validation/test splits defined in Appendix~\ref{app:benchmark-details}. Our experiments comprise static skill-value baselines (no skill, handcrafted, and generated), a single refinement pass, and skill evolution; for evolved skills we measure transfer along the role and model axes and report token efficiency. Cross-role transfer uses the three high-overlap skills shared by four roles each---\texttt{pdf}, \texttt{xlsx}, and \texttt{validation}.

\paragraph{Models.} The baseline evaluation runs a panel of open- and closed-weight models, grouped by family and size tier (S/M/L): the GPT~5.4 family (GPT~5.4, GPT~5.4 Mini), GPT-oss (120B, 20B), the Qwen~3.5 family (397B-FP8, 122B-A10B, 35B-A3B, 9B), the Gemma~4 family (31B, 26B-A4B, E4B), DeepSeek~V4 Flash, and Nemotron~3 (120B). We also use a harness around the following models to rewrite skills during evolution: Claude Sonnet~4.6 (Claude Code), GPT~5.5 (Codex), and DeepSeek~V4 Flash (Hermes). All of our experiments use the models listed above; for the cross-model transfer experiment (Figure~\ref{fig:model-experience}) we additionally add Qwen 3.5-27B and Llama 8B.

\paragraph{Splits.} Tasks are assigned to three splits: \texttt{train} (used for evolution), \texttt{test} (unseen tasks for every role+skill combination), and \texttt{valdation}.



\section{Token Usage for auto-agents}

We measure the inference cost of procedural memory in an agentic loop: Claude (Sonnet~4.6) and the small open backbone Hermes (Qwen~3.5-35B-A3B) run three
representative auto-agent tasks, and we record generated tokens, total tokens,
and dollar cost under four skill scenarios --- no skill, handcrafted, evolved, and
self-evolved (Table~\ref{tab:claude-hermes-usage}, means over four runs).
Evolved skills generally cut total token usage relative to the handcrafted
baseline (e.g.\ 521k$\rightarrow$195k for Claude on Kafka Lag Anomaly Detection),
lowering cost without sacrificing task success.

\begin{table}[h!]
\centering
\scriptsize
\setlength{\tabcolsep}{3pt}
\resizebox{\columnwidth}{!}{%
\begin{tabular}{@{}ll|rr|rr|rr@{}}
\toprule
& & \multicolumn{2}{c|}{\textbf{GH Actions Opt.}} & \multicolumn{2}{c|}{\textbf{Kafka Lag}} & \multicolumn{2}{c}{\textbf{PPTX Fmt.}} \\
\textbf{Metric} & \textbf{Scenario} & Claude & Hermes & Claude & Hermes & Claude & Hermes \\
\midrule
\multirow{4}{*}{\textbf{Gen (k)}}
 & No Skill      & 25.4 & 11.7 & 8.5 & 14.7 & 18.3 & 27.2 \\
 & Handcrafted   & 26.2 & 27.8 & 8.3 & 6.2 & 18.4 & 11.2 \\
 & Evolved       & 21.6 & 27.1 & 2.1 & 3.2 & 10.3 & 18.7 \\
 & Self-Evolved  & 9.6 & 19.4 & 12.7 & 3.1 & 10.6 & 22.9 \\
\midrule
\multirow{4}{*}{\textbf{Total (k)}}
 & No Skill      & 505.3 & 278.2 & 421.2 & 406.5 & 472.9 & 1526.8 \\
 & Handcrafted   & 625.0 & 460.8 & 521.1 & 302.5 & 664.9 & 335.5 \\
 & Evolved       & 285.8 & 308.8 & 194.7 & 254.4 & 504.8 & 476.5 \\
 & Self-Evolved  & 227.4 & 358.7 & 738.8 & 228.6 & 352.2 & 1419.3 \\
\midrule
\multirow{4}{*}{\textbf{Cost (\$)}}
 & No Skill      & 0.675 & 0.049 & 0.333 & 0.070 & 0.535 & 0.237 \\
 & Handcrafted   & 0.728 & 0.088 & 0.388 & 0.048 & 0.617 & 0.057 \\
 & Evolved       & 0.538 & 0.067 & 0.150 & 0.038 & 0.422 & 0.083 \\
 & Self-Evolved  & 0.306 & 0.067 & 0.519 & 0.035 & 0.371 & 0.218 \\
\bottomrule
\end{tabular}
}%
\caption{Token usage and cost: Claude (Sonnet\,4.6) vs.\ Hermes (Qwen\,3.5-35B-A3B) on three auto-agent tasks across four skill scenarios. Gen\,(k)\,=\,generated tokens\,/\,1000; Total\,(k)\,=\,total tokens\,/\,1000; Cost in USD. Means over 4 runs.}
\label{tab:claude-hermes-usage}
\end{table}

\section{Reflector Ablation}
To show that skill evolution improves task performance independently of the reasoner that performs it, we evolve the \texttt{pptx} and \texttt{xlsx} skills with four reasoners --- the Claude, Hermes, and Codex agents, plus our own script with no agent involved --- while holding the solver fixed at GPT-oss-120B. Each reasoner tunes a handcrafted baseline on a train subset of $n=1$ up to $n=5$ tasks and is tested on 3 held-out tasks. As Table~\ref{tab:reasoner-ablation-m1} shows, diverse training beats narrow for every reasoner, confirming the gain comes from evolution itself rather than from any particular agent.

\definecolor{rowoss}{HTML}{F3EEFB}
\definecolor{rowclaude}{HTML}{FDF4EC}
\definecolor{rowcodex}{HTML}{F0F8F2}
\definecolor{rowhermes}{HTML}{FCF0F5}
\begin{table}[h!]
\centering
\scriptsize
\setlength{\tabcolsep}{4pt}
\begin{tabular}{@{}l|rr@{}}
\toprule
\textbf{Reasoner} & \textbf{Narrow} & \textbf{Diverse} \\
\midrule
\multicolumn{3}{@{}l}{\textit{pptx}} \\[1pt]
\rowcolor{rowoss}Script (GPT-oss-120B) & 60.1 & \textbf{79.3} \\
\rowcolor{rowclaude}Claude Code (Opus\,4.8) & 56.8 & \textbf{87.5} \\
\rowcolor{rowcodex}Codex (GPT-5.5) & 65.0 & \textbf{82.7} \\
\rowcolor{rowhermes}Hermes (DeepSeek\,V4\,Flash) & 59.1 & \textbf{82.7} \\
\midrule
\multicolumn{3}{@{}l}{\textit{xlsx}} \\[1pt]
\rowcolor{rowoss}Script (GPT-oss-120B) & 62.6 & \textbf{68.1} \\
\rowcolor{rowclaude}Claude Code (Opus\,4.8) & 51.7 & \textbf{68.2} \\
\rowcolor{rowcodex}Codex (GPT-5.5) & 45.8 & \textbf{56.1} \\
\rowcolor{rowhermes}Hermes (DeepSeek\,V4\,Flash) & 53.5 & \textbf{67.4} \\
\bottomrule
\end{tabular}
\caption{Reasoner ablation --- M1 (test, \%); best result per setting. Bold marks the larger value in each row.}
\label{tab:reasoner-ablation-m1}
\end{table}

\onecolumn
\section{Train Split Results}
\begin{table*}[ht!]
\centering
\scriptsize
\setlength{\tabcolsep}{1.5pt}
\begin{tabular}{@{}ll|ccc|ccc|ccc|ccc|ccc|ccc|cccc@{}}
\toprule
& & \multicolumn{3}{c|}{\textbf{DE}} & \multicolumn{3}{c|}{\textbf{DS}} & \multicolumn{3}{c|}{\textbf{GenAI}} & \multicolumn{3}{c|}{\textbf{Infra}} & \multicolumn{3}{c|}{\textbf{PM}} & \multicolumn{3}{c|}{\textbf{SWE}} & \multicolumn{4}{c}{\textbf{Aggregate}} \\
\textbf{Model} & \textbf{Size} & $\varnothing$ & H & G & $\varnothing$ & H & G & $\varnothing$ & H & G & $\varnothing$ & H & G & $\varnothing$ & H & G & $\varnothing$ & H & G & $\varnothing$ & H & G & $\Delta_{best-no}$ \\
\midrule
Gemma 4 31B & M & --- & --- & --- & --- & --- & --- & --- & --- & --- & --- & --- & --- & --- & --- & --- & --- & --- & --- & --- & --- & --- & --- \\
Gemma 4 26B A4B & M & 27.4 & \cellcolor{lowgain}28.1 & \cellcolor{medgain}\textbf{29.5} & 25.5 & \cellcolor{reggain}25.4 & \cellcolor{reggain}25.3 & 29.6 & \cellcolor{medgain}31.6 & \cellcolor{medgain}32.5 & 41.4 & \cellcolor{highgain}45.8 & \cellcolor{highgain}\textbf{49.3} & 25.7 & \cellcolor{medgain}28.4 & \cellcolor{highgain}29.1 & 24.9 & \cellcolor{reggain}2.4 & \cellcolor{medgain}\textbf{26.5} & 28.9 & \cellcolor{lowgain}30.0 & \cellcolor{medgain}\textbf{31.7} & \cellcolor{best}{\textbf{+2.8}} \\
Gemma 4 E4B & S & 14.7 & \cellcolor{reggain}14.5 & \cellcolor{lowgain}15.6 & 21.6 & \cellcolor{lowgain}22.2 & \cellcolor{reggain}20.2 & 21.1 & \cellcolor{medgain}23.7 & \cellcolor{medgain}23.8 & 37.7 & \cellcolor{lowgain}38.1 & \cellcolor{reggain}36.2 & 19.3 & \cellcolor{lowgain}19.4 & \cellcolor{reggain}17.9 & 19.9 & \cellcolor{reggain}19.0 & \cellcolor{lowgain}21.4 & 22.1 & \cellcolor{lowgain}22.6 & \cellcolor{lowgain}22.4 & \cellcolor{lowgain}{+0.5} \\
\midrule
Qwen 3.5-397B-FP8 & L & --- & --- & --- & --- & --- & --- & --- & --- & --- & --- & --- & --- & --- & --- & --- & --- & --- & --- & --- & --- & --- & --- \\
Qwen 3.5-122B-A10B & L & 22.9 & \cellcolor{lowgain}23.4 & \cellcolor{medgain}24.8 & 25.1 & \cellcolor{lowgain}25.9 & \cellcolor{reggain}24.0 & 28.4 & \cellcolor{lowgain}29.8 & \cellcolor{lowgain}29.6 & 42.4 & \cellcolor{medgain}44.6 & \cellcolor{highgain}46.2 & 24.9 & \cellcolor{highgain}29.2 & \cellcolor{medgain}27.2 & 20.3 & \cellcolor{lowgain}21.7 & \cellcolor{lowgain}21.3 & 27.0 & \cellcolor{medgain}28.7 & \cellcolor{lowgain}28.4 & \cellcolor{medgain}{+1.7} \\
Qwen 3.5-35B-A3B & M & 21.0 & \cellcolor{reggain}20.9 & \cellcolor{reggain}20.6 & 23.2 & \cellcolor{reggain}22.6 & \cellcolor{reggain}0.0 & 26.3 & \cellcolor{reggain}25.6 & \cellcolor{reggain}25.5 & 36.8 & \cellcolor{medgain}38.7 & \cellcolor{highgain}42.0 & 22.9 & \cellcolor{highgain}27.1 & \cellcolor{medgain}25.5 & 18.0 & \cellcolor{lowgain}19.1 & \cellcolor{medgain}20.7 & 24.4 & \cellcolor{lowgain}25.6 & \cellcolor{medgain}26.3 & \cellcolor{medgain}{+1.9} \\
Qwen 3.5-9B & S & 13.3 & \cellcolor{lowgain}14.6 & \cellcolor{lowgain}14.1 & 16.6 & \cellcolor{lowgain}17.0 & \cellcolor{lowgain}16.7 & 14.1 & \cellcolor{lowgain}15.5 & \cellcolor{highgain}17.7 & 37.8 & \cellcolor{reggain}32.5 & \cellcolor{reggain}35.6 & 14.3 & \cellcolor{lowgain}15.5 & \cellcolor{lowgain}15.1 & 17.0 & \cellcolor{lowgain}17.2 & \cellcolor{reggain}16.2 & 18.6 & \cellcolor{reggain}18.5 & \cellcolor{lowgain}19.0 & \cellcolor{lowgain}{+0.4} \\
\midrule
GPT-oss-120B & L & 24.1 & \cellcolor{lowgain}24.9 & \cellcolor{lowgain}24.2 & \textbf{27.6} & \cellcolor{reggain}26.5 & \cellcolor{reggain}27.3 & 32.2 & \cellcolor{medgain}33.8 & \cellcolor{medgain}\textbf{34.1} & 43.3 & \cellcolor{lowgain}43.9 & \cellcolor{lowgain}43.6 & 28.9 & \cellcolor{highgain}32.3 & \cellcolor{highgain}\textbf{32.5} & 22.7 & \cellcolor{lowgain}22.9 & \cellcolor{lowgain}24.1 & 29.4 & \cellcolor{lowgain}30.3 & \cellcolor{lowgain}30.5 & \cellcolor{lowgain}{+1.1} \\
GPT-oss-20B & M & 24.1 & \cellcolor{reggain}23.4 & \cellcolor{lowgain}24.6 & 24.5 & \cellcolor{reggain}23.0 & \cellcolor{reggain}23.6 & 28.3 & \cellcolor{lowgain}29.1 & \cellcolor{medgain}31.0 & 42.1 & \cellcolor{lowgain}42.3 & \cellcolor{lowgain}42.8 & 25.1 & \cellcolor{medgain}27.3 & \cellcolor{highgain}28.7 & 20.6 & \cellcolor{reggain}20.6 & \cellcolor{lowgain}21.8 & 27.1 & \cellcolor{lowgain}27.2 & \cellcolor{lowgain}28.3 & \cellcolor{lowgain}{+1.2} \\
\bottomrule
\end{tabular}
\captionsetup{width=\linewidth}
\caption{Static benchmark performance on \benchname{}: average task M1 metric per role under three skill conditions ($\varnothing$ = no skill, H = handcrafted, G = LLM-generated). Evaluated on the 185-task train split. Per-task metric is the mean over attempts of $\mathrm{tests\_passed}/\mathrm{task\_total}$; per-role and aggregate values are unweighted means across tasks. Colors mark the gain from skills: \colorbox{best}{best}, \colorbox{highgain}{$>3$}, \colorbox{medgain}{$1.5$--$3$}, \colorbox{lowgain}{$0$--$1.5$}, \colorbox{reggain}{$<0$}.}
\label{tab:pass-rate-train-subset}
\end{table*}

\begin{table*}[ht!]
\centering
\scriptsize
\setlength{\tabcolsep}{1.5pt}
\begin{tabular}{@{}ll|ccc|ccc|ccc|ccc|ccc|ccc|cccc@{}}
\toprule
& & \multicolumn{3}{c|}{\textbf{DE}} & \multicolumn{3}{c|}{\textbf{DS}} & \multicolumn{3}{c|}{\textbf{GenAI}} & \multicolumn{3}{c|}{\textbf{Infra}} & \multicolumn{3}{c|}{\textbf{PM}} & \multicolumn{3}{c|}{\textbf{SWE}} & \multicolumn{4}{c}{\textbf{Aggregate}} \\
\textbf{Model} & \textbf{Size} & $\varnothing$ & H & G & $\varnothing$ & H & G & $\varnothing$ & H & G & $\varnothing$ & H & G & $\varnothing$ & H & G & $\varnothing$ & H & G & $\varnothing$ & H & G & $\Delta_{best-no}$ \\
\midrule
Gemma 4 31B & M & --- & --- & --- & --- & --- & --- & --- & --- & --- & --- & --- & --- & --- & --- & --- & --- & --- & --- & --- & --- & --- & --- \\
Gemma 4 26B A4B & M & 15.0 & \cellcolor{lowgain}16.2 & \cellcolor{medgain}\textbf{16.7} & \textbf{16.7} & \cellcolor{reggain}12.6 & \cellcolor{reggain}13.6 & 19.4 & \cellcolor{medgain}20.9 & \cellcolor{medgain}\textbf{21.1} & 30.0 & \cellcolor{highgain}38.8 & \cellcolor{highgain}\textbf{42.7} & 16.0 & \cellcolor{reggain}15.6 & \cellcolor{lowgain}17.0 & 11.6 & \cellcolor{reggain}0.0 & \cellcolor{highgain}\textbf{15.2} & 17.9 & \cellcolor{medgain}19.4 & \cellcolor{medgain}\textbf{20.7} & \cellcolor{medgain}{+2.8} \\
Gemma 4 E4B & S & 6.7 & \cellcolor{reggain}6.1 & \cellcolor{reggain}6.6 & 9.5 & \cellcolor{lowgain}10.0 & \cellcolor{reggain}8.6 & 8.6 & \cellcolor{reggain}8.0 & \cellcolor{reggain}6.2 & 16.8 & \cellcolor{medgain}18.8 & \cellcolor{highgain}23.2 & 7.0 & \cellcolor{reggain}6.4 & \cellcolor{lowgain}8.2 & 6.9 & \cellcolor{reggain}6.2 & \cellcolor{highgain}11.8 & 9.2 & \cellcolor{reggain}9.1 & \cellcolor{lowgain}10.6 & \cellcolor{lowgain}{+1.4} \\
\midrule
Qwen 3.5-397B-FP8 & L & --- & --- & --- & --- & --- & --- & --- & --- & --- & --- & --- & --- & --- & --- & --- & --- & --- & --- & --- & --- & --- & --- \\
Qwen 3.5-122B-A10B & L & 12.7 & \cellcolor{lowgain}14.1 & \cellcolor{lowgain}13.1 & 12.1 & \cellcolor{lowgain}12.7 & \cellcolor{reggain}11.2 & 15.5 & \cellcolor{medgain}17.2 & \cellcolor{lowgain}15.5 & 36.1 & \cellcolor{medgain}38.2 & \cellcolor{highgain}40.4 & 11.4 & \cellcolor{highgain}17.6 & \cellcolor{highgain}16.8 & 7.5 & \cellcolor{lowgain}8.4 & \cellcolor{highgain}11.8 & 15.5 & \cellcolor{medgain}17.5 & \cellcolor{medgain}17.6 & \cellcolor{medgain}{+2.1} \\
Qwen 3.5-35B-A3B & M & 9.5 & \cellcolor{reggain}6.4 & \cellcolor{reggain}9.4 & 9.8 & \cellcolor{reggain}9.7 & \cellcolor{reggain}0.0 & 14.1 & \cellcolor{reggain}13.9 & \cellcolor{lowgain}14.2 & 28.4 & \cellcolor{highgain}32.0 & \cellcolor{highgain}36.6 & 9.2 & \cellcolor{highgain}13.4 & \cellcolor{medgain}12.0 & 7.4 & \cellcolor{lowgain}8.1 & \cellcolor{medgain}10.0 & 12.8 & \cellcolor{lowgain}14.0 & \cellcolor{highgain}15.9 & \cellcolor{best}{\textbf{+3.1}} \\
Qwen 3.5-9B & S & 5.9 & \cellcolor{lowgain}5.9 & \cellcolor{reggain}5.8 & 7.7 & \cellcolor{reggain}7.0 & \cellcolor{reggain}6.4 & 2.8 & \cellcolor{medgain}4.5 & \cellcolor{medgain}5.6 & 27.7 & \cellcolor{reggain}24.8 & \cellcolor{lowgain}28.9 & 2.8 & \cellcolor{medgain}5.4 & \cellcolor{medgain}5.6 & 5.7 & \cellcolor{lowgain}5.9 & \cellcolor{medgain}8.1 & 8.6 & \cellcolor{lowgain}8.7 & \cellcolor{lowgain}9.8 & \cellcolor{lowgain}{+1.2} \\
\midrule
GPT-oss-120B & L & 12.5 & \cellcolor{reggain}12.0 & \cellcolor{reggain}11.7 & 14.4 & \cellcolor{reggain}14.1 & \cellcolor{reggain}13.5 & 19.4 & \cellcolor{medgain}\textbf{21.7} & \cellcolor{lowgain}20.6 & 31.4 & \cellcolor{medgain}34.3 & \cellcolor{highgain}36.4 & 15.8 & \cellcolor{highgain}20.6 & \cellcolor{highgain}\textbf{21.0} & 9.9 & \cellcolor{reggain}9.4 & \cellcolor{highgain}13.7 & 16.9 & \cellcolor{lowgain}18.2 & \cellcolor{medgain}19.0 & \cellcolor{medgain}{+2.1} \\
GPT-oss-20B & M & 13.0 & \cellcolor{reggain}10.3 & \cellcolor{reggain}12.0 & 11.2 & \cellcolor{reggain}10.9 & \cellcolor{reggain}11.1 & 16.9 & \cellcolor{reggain}16.1 & \cellcolor{medgain}19.1 & 28.0 & \cellcolor{lowgain}28.4 & \cellcolor{highgain}32.9 & 14.2 & \cellcolor{lowgain}14.6 & \cellcolor{medgain}16.8 & 7.1 & \cellcolor{lowgain}7.9 & \cellcolor{highgain}11.2 & 14.7 & \cellcolor{reggain}14.3 & \cellcolor{medgain}16.7 & \cellcolor{medgain}{+2.0} \\
\bottomrule
\end{tabular}
\captionsetup{width=\linewidth}
\caption{Static benchmark performance on \benchname{}: M2 metric per role under three skill conditions ($\varnothing$ = no skill, H = handcrafted, G = LLM-generated). Evaluated on the 185-task train split. Per-task metric is the fraction of attempts where $\mathrm{tests\_passed} = \mathrm{task\_total}$; per-role and aggregate values are unweighted means across tasks. Colors mark the gain from skills: \colorbox{best}{best}, \colorbox{highgain}{$>3$}, \colorbox{medgain}{$1.5$--$3$}, \colorbox{lowgain}{$0$--$1.5$}, \colorbox{reggain}{$<0$}.}
\label{tab:perfect-rate-train-subset}
\end{table*}

\newpage
\section{Full Results}
\label{app:test-results}
\definecolor{highgain}{HTML}{C8E6C9}
\definecolor{medgain}{HTML}{E8F5E9}
\definecolor{lowgain}{HTML}{FFF8E1}
\definecolor{best}{HTML}{A5D6A7}
\definecolor{reggain}{HTML}{FFCDD2}

\begin{table*}[ht!]
\centering
\scriptsize
\setlength{\tabcolsep}{1.5pt}
\begin{tabular}{@{}ll|ccc|ccc|ccc|ccc|ccc|ccc|cccc@{}}
\toprule
& & \multicolumn{3}{c|}{\textbf{DE}} & \multicolumn{3}{c|}{\textbf{DS}} & \multicolumn{3}{c|}{\textbf{GenAI}} & \multicolumn{3}{c|}{\textbf{Infra}} & \multicolumn{3}{c|}{\textbf{PM}} & \multicolumn{3}{c|}{\textbf{SWE}} & \multicolumn{4}{c}{\textbf{Aggregate}} \\
\textbf{Model} & \textbf{Size} & $\varnothing$ & H & G & $\varnothing$ & H & G & $\varnothing$ & H & G & $\varnothing$ & H & G & $\varnothing$ & H & G & $\varnothing$ & H & G & $\varnothing$ & H & G & $\Delta_{best-no}$ \\
\midrule
GPT 5.4 & L & \textbf{73.3} & \cellcolor{reggain}70.8 & \cellcolor{reggain}72.5 & 42.0 & \cellcolor{highgain}47.0 & \cellcolor{highgain}\textbf{55.0} & 40.0 & \cellcolor{highgain}\textbf{43.1} & \cellcolor{lowgain}40.0 & 50.0 & \cellcolor{reggain}48.7 & \cellcolor{highgain}56.3 & 38.5 & \cellcolor{highgain}\textbf{44.6} & \cellcolor{reggain}33.8 & 32.6 & \cellcolor{lowgain}32.6 & \cellcolor{lowgain}32.6 & 47.6 & \cellcolor{lowgain}49.0 & \cellcolor{medgain}\textbf{50.1} & \cellcolor{lowgain}{+2.5} \\
GPT 5.4 Mini 4 & M & 63.3 & \cellcolor{highgain}66.7 & \cellcolor{lowgain}64.2 & 43.0 & \cellcolor{reggain}33.0 & \cellcolor{highgain}48.0 & 35.8 & \cellcolor{reggain}31.6 & \cellcolor{highgain}45.3 & \textbf{60.0} & \cellcolor{reggain}43.7 & \cellcolor{reggain}48.7 & 27.7 & \cellcolor{reggain}24.6 & \cellcolor{highgain}30.8 & 26.3 & \cellcolor{lowgain}26.3 & \cellcolor{reggain}23.2 & 44.0 & \cellcolor{reggain}39.5 & \cellcolor{lowgain}44.9  & \cellcolor{lowgain}{+0.9} \\
DeepSeek V4 Flash & L & 56.8 & \cellcolor{lowgain}56.8 & \cellcolor{medgain}59.2 & 28.7 & \cellcolor{lowgain}29.6 & \cellcolor{highgain}33.9 & 31.6 & \cellcolor{reggain}30.5 & \cellcolor{highgain}35.8 & 48.8 & \cellcolor{reggain}46.3 & \cellcolor{reggain}41.2 & 11.4 & \cellcolor{highgain}17.1 & \cellcolor{highgain}18.1 & 29.5 & \cellcolor{highgain}32.6 & \cellcolor{lowgain}30.5 & 34.6 & \cellcolor{lowgain}35.8 & \cellcolor{medgain}37.1  & \cellcolor{lowgain}{+2.5}\\
Nemotron 3 120B & M & 44.8 & \cellcolor{reggain}29.6 & \cellcolor{reggain}40.8 & 29.6 & \cellcolor{medgain}31.3 & \cellcolor{reggain}28.7 & 42.1 & \cellcolor{reggain}36.8 & \cellcolor{reggain}35.8 & 37.5 & \cellcolor{reggain}31.2 & \cellcolor{medgain}41.3 & 14.3 & \cellcolor{highgain}21.9 & \cellcolor{highgain}24.8 & 22.1 &\cellcolor{reggain}17.9 & \cellcolor{highgain}27.4 & 31.9 & \cellcolor{reggain}28.1 & \cellcolor{lowgain}33.0  & \cellcolor{lowgain}{+1.1} \\
\midrule
Gemma 4 31B & M & 51.3 & \cellcolor{lowgain}52.5 & \cellcolor{highgain}57.9 & 45.5 & \cellcolor{reggain}45.0 & \cellcolor{highgain}49.5 & 35.3 & \cellcolor{lowgain}35.3 & \cellcolor{reggain}33.7 & 43.1 & \cellcolor{medgain}45.6 & \cellcolor{lowgain}44.4 & 10.0 & \cellcolor{highgain}13.9 & \cellcolor{highgain}23.9 & \textbf{33.7} & \cellcolor{reggain}27.4 & \cellcolor{reggain}28.4 & 38.5 & \cellcolor{reggain}38.4 & \cellcolor{medgain}41.3  & \cellcolor{lowgain}{+2.8} \\
Gemma 4 26B A4B & M & 54.6 & \cellcolor{highgain}60.4 & \cellcolor{reggain}53.3 & 37.5 & \cellcolor{highgain}42.0 & \cellcolor{highgain}48.0 & 37.4 & \cellcolor{medgain}40.0 & \cellcolor{medgain}40.0 & 34.4 & \cellcolor{reggain}28.1 & \cellcolor{highgain}40.0 & 16.9 & \cellcolor{highgain}24.6 & \cellcolor{medgain}20.0 & 25.3 & \cellcolor{lowgain}25.8 & \cellcolor{lowgain}26.8 & 36.2 & \cellcolor{medgain}{38.8} & \cellcolor{highgain}{39.7} & \cellcolor{highgain}{+3.5} \\
Gemma 4 E4B & S & 28.3 & \cellcolor{medgain}31.2 & \cellcolor{highgain}35.0 & 20.5 & \cellcolor{reggain}18.5 & \cellcolor{highgain}24.5 & 15.3 & \cellcolor{highgain}23.7 & \cellcolor{highgain}29.5 & 13.8 & \cellcolor{lowgain}14.4 & \cellcolor{highgain}21.2 & 8.5 & \cellcolor{medgain}11.5 & \cellcolor{highgain}13.1 & 23.2 & \cellcolor{reggain}17.9 & \cellcolor{reggain}13.7 & 19.4 & \cellcolor{lowgain}{20.6} & \cellcolor{highgain}{24.0} & \cellcolor{highgain}{+4.6} \\
\midrule
Qwen 3.5-397B-FP8 & L & 55.4 & \cellcolor{highgain}59.2 & \cellcolor{highgain}62.1  & 36.0  & \cellcolor{reggain}35.2 & \cellcolor{highgain}40.2 & 35.3 & \cellcolor{lowgain}35.5 & \cellcolor{medgain}38.7 & 48.1 & \cellcolor{reggain}46.9 & \cellcolor{reggain}45.0 & 12.3 & \cellcolor{lowgain}13.5 & \cellcolor{highgain}16.9 & 28.7 & \cellcolor{medgain}31.3 & \cellcolor{highgain}32.4 & 37.8 & \cellcolor{lowgain}38.9 & \cellcolor{highgain}41.3 & \cellcolor{medgain}{+3.5}\\
Qwen 3.5-122B-A10B & L & 55.4 &\cellcolor{reggain} 52.9 & \cellcolor{medgain}57.5 & 36.0  & \cellcolor{reggain}33.5 & \cellcolor{highgain}41.0 & 30.0 & \cellcolor{highgain}34.2 & \cellcolor{highgain}35.3 & 44.4 & \cellcolor{lowgain}45.0 & \cellcolor{reggain}42.5 & 12.3 & \cellcolor{highgain}19.2 & \cellcolor{highgain}20.8 & 29.5 & \cellcolor{lowgain}29.5 & \cellcolor{lowgain}31.1 & 36.5 & \cellcolor{lowgain}37.1 & \cellcolor{highgain}39.7 & \cellcolor{lowgain}{+3.2} \\
Qwen 3.5-35B-A3B & M & 37.9 & \cellcolor{highgain}43.3 & \cellcolor{highgain}43.8 & 21.5 & \cellcolor{highgain}25.5 & \cellcolor{highgain}{30.5} & 27.9 & \cellcolor{highgain}31.6 & \cellcolor{highgain}{36.9} & 28.1 & \cellcolor{reggain}23.7 & \cellcolor{highgain}35.0 & 13.1 & \cellcolor{lowgain}13.1 & \cellcolor{lowgain}14.6 & 26.3 & \cellcolor{reggain}20.0 & \cellcolor{reggain}24.2 & 26.9 & \cellcolor{lowgain}27.7 & \cellcolor{highgain}32.2 & \cellcolor{best}{\textbf{+5.3}} \\
Qwen 3.5-9B & S & 23.7 & \cellcolor{reggain}22.1 & \cellcolor{medgain}26.7 & 12.5 & \cellcolor{reggain}11.5 & \cellcolor{highgain}17.0 & 17.9 & \cellcolor{reggain}15.3 & \cellcolor{lowgain}18.9 & 11.3 & \cellcolor{medgain}13.8 & \cellcolor{best}20.6 & 6.2 & \cellcolor{highgain}10.8 & \cellcolor{highgain}14.6 & 16.8 & \cellcolor{reggain}13.2 & \cellcolor{reggain}11.6 & 15.7 & \cellcolor{reggain}15.0 & \cellcolor{medgain}18.7 & \cellcolor{medgain}{+3.0} \\
\midrule
GPT-oss-120B & L & 54.2 & \cellcolor{lowgain}55.0 & \cellcolor{lowgain}55.4 & 44.5 & \cellcolor{medgain}47.0 & \cellcolor{highgain}49.0 & 38.3 & \cellcolor{highgain}42.6 & \cellcolor{highgain}41.6 & 57.5 & \cellcolor{reggain}51.3 & \cellcolor{medgain}58.7 & 32.3 & \cellcolor{reggain}30.8 & \cellcolor{lowgain}33.1 & 30.0 & \cellcolor{reggain}{29.5} & \cellcolor{lowgain}31.0 & 43.2 & \cellcolor{lowgain}43.7 & \cellcolor{medgain}45.6 & \cellcolor{medgain}{+2.4} \\
GPT-oss-20B & M & 42.9 & \cellcolor{reggain}{41.2} & \cellcolor{lowgain}42.9 & 30.0 & \cellcolor{medgain}32.0 & \cellcolor{reggain}29.0 & 32.6 & \cellcolor{lowgain}32.6 & \cellcolor{lowgain}32.6 & 30.6 & \cellcolor{reggain}25.6 & \cellcolor{reggain}30.0 & 16.1 & \cellcolor{highgain}22.3 & \cellcolor{lowgain}16.9 & 25.3 & \cellcolor{medgain}{27.4} &\cellcolor{reggain}{24.7} & 30.9 & \cellcolor{lowgain}31.3 & \cellcolor{reggain}{30.6} & \cellcolor{lowgain}{+0.4} \\
\bottomrule
\end{tabular}
\caption{Static benchmark performance on \benchname{}: (M1,\%) per role under three skill conditions ($\varnothing$ = no skill, H = handcrafted, G = LLM-generated). Aggregate columns show average full-pass rate across all tasks and best delta between no-skill prompt and best result using a skill. Colors mark the gain from skills: \colorbox{best}{best}, \colorbox{highgain}{>3}, \colorbox{medgain}{1.5..3}, \colorbox{lowgain}{0..1.5} \colorbox{reggain}{<0}.}
\label{tab:pass-rate-test-subset}
\end{table*}

\begin{table*}[h!]
\centering
\scriptsize
\setlength{\tabcolsep}{1.5pt}
\begin{tabular}{@{}ll|ccc|ccc|ccc|ccc|ccc|ccc|cccc@{}}
\toprule
& & \multicolumn{3}{c|}{\textbf{DE}} & \multicolumn{3}{c|}{\textbf{DS}} & \multicolumn{3}{c|}{\textbf{GenAI}} & \multicolumn{3}{c|}{\textbf{Infra}} & \multicolumn{3}{c|}{\textbf{PM}} & \multicolumn{3}{c|}{\textbf{SWE}} & \multicolumn{4}{c}{\textbf{Aggregate}} \\
\textbf{Model} & \textbf{Size} & $\varnothing$ & H & G & $\varnothing$ & H & G & $\varnothing$ & H & G & $\varnothing$ & H & G & $\varnothing$ & H & G & $\varnothing$ & H & G & $\varnothing$ & H & G & $\Delta_{best-no}$ \\
\midrule
GPT 5.4 & L & \textbf{90.3} & \cellcolor{reggain}90.2 & \cellcolor{reggain}89.7 & 79.7 & \cellcolor{lowgain}79.8 & \cellcolor{medgain}\textbf{82.3} & 70.3 & \cellcolor{lowgain}\textbf{71.6} & \cellcolor{reggain}68.1 & 79.2 & \cellcolor{medgain}\textbf{81.1} & \cellcolor{lowgain}80.2 & 73.2 & \cellcolor{lowgain}\textbf{73.2} & \cellcolor{reggain}66.0 & \textbf{91.5} & \cellcolor{reggain}76.0 & \cellcolor{reggain}72.1 & \textbf{81.6} & \cellcolor{reggain}79.4 & \cellcolor{reggain}77.5 & \cellcolor{reggain}{-2.1} \\
GPT 5.4 Mini 4 & M & 82.1 & \cellcolor{medgain}84.4 & \cellcolor{reggain}82.0 & 73.1 & \cellcolor{reggain}67.7 & \cellcolor{reggain}72.5 & 63.8 & \cellcolor{medgain}66.2 & \cellcolor{highgain}67.8 & 72.6 & \cellcolor{reggain}64.9 & \cellcolor{reggain}67.7 & 60.6 & \cellcolor{reggain}58.7 & \cellcolor{highgain}\textbf{67.1} & 52.6 & \cellcolor{highgain}55.7 & \cellcolor{reggain}50.0 & 68.4 & \cellcolor{reggain}67.6 & \cellcolor{lowgain}68.6 & \cellcolor{lowgain}{+0.2} \\
DeepSeek V4 Flash & L & 75.1 & \cellcolor{lowgain}75.8 & \cellcolor{highgain}78.5 & 52.8 & \cellcolor{reggain}50.6 & \cellcolor{lowgain}53.2 & 60.4 & \cellcolor{reggain}51.2 & \cellcolor{reggain}56.9 & 66.7 & \cellcolor{lowgain}67.5 & \cellcolor{reggain}63.3 & 41.2 & \cellcolor{highgain}44.6 & \cellcolor{medgain}43.5 & 56.7 & \cellcolor{highgain}62.6 & \cellcolor{medgain}59.6 & 58.9 & \cellcolor{reggain}58.8 & \cellcolor{lowgain}59.6 & \cellcolor{lowgain}{+0.6} \\
Nemotron 3 120B & M & 69.2 & \cellcolor{reggain}48.7 & \cellcolor{reggain}57.8 & 47.1 & \cellcolor{lowgain}48.6 & \cellcolor{medgain}48.9 & 62.6 & \cellcolor{reggain}55.6 & \cellcolor{reggain}51.4 & 64.7 & \cellcolor{reggain}61.8 & \cellcolor{reggain}63.3 & 37.5 & \cellcolor{medgain}40.1 & \cellcolor{highgain}44.0 & 49.2 & \cellcolor{reggain}42.2 & \cellcolor{highgain}54.6 & 55.0 & \cellcolor{reggain}49.0 & \cellcolor{reggain}53.0 & \cellcolor{reggain}{-2.0} \\
\midrule
Gemma 4 31B & M & 69.1 & \cellcolor{medgain}71.3 & \cellcolor{highgain}77.9 & 71.4 & \cellcolor{reggain}65.3 & \cellcolor{lowgain}72.6 & 62.7 & \cellcolor{lowgain}63.0 & \cellcolor{medgain}65.0 & 59.9 & \cellcolor{highgain}64.3 & \cellcolor{highgain}66.6 & 50.6 & \cellcolor{reggain}50.5 & \cellcolor{highgain}56.2 & 59.1 & \cellcolor{reggain}51.1 & \cellcolor{reggain}51.7 & 63.2 & \cellcolor{reggain}61.9 & \cellcolor{medgain}66.1 & \cellcolor{medgain}{+2.8} \\
Gemma 4 26B A4B & M & 81.0 & \cellcolor{reggain}77.5 & \cellcolor{reggain}75.6 & 58.8 & \cellcolor{highgain}69.9 & \cellcolor{highgain}72.7 & 63.9 & \cellcolor{lowgain}64.8 & \cellcolor{reggain}61.0 & 53.4 & \cellcolor{highgain}64.4 & \cellcolor{highgain}66.7 & 46.8 & \cellcolor{highgain}56.1 & \cellcolor{highgain}52.9 & 56.3 & \cellcolor{reggain}55.1 & \cellcolor{reggain}51.6 & 61.7 & \cellcolor{highgain}65.8 & \cellcolor{medgain}64.4 & \cellcolor{best}{\textbf{+4.1}} \\
Gemma 4 E4B & S & 43.3 & \cellcolor{reggain}37.5 & \cellcolor{reggain}40.8 & 42.5 & \cellcolor{reggain}40.6 & \cellcolor{reggain}41.8 & 41.4 & \cellcolor{highgain}50.0 & \cellcolor{highgain}50.1 & 42.1 & \cellcolor{highgain}46.3 & \cellcolor{highgain}45.5 & 34.2 & \cellcolor{lowgain}34.2 & \cellcolor{medgain}36.7 & 47.7 & \cellcolor{reggain}41.2 & \cellcolor{reggain}38.1 & 42.3 & \cellcolor{reggain}41.7 & \cellcolor{reggain}42.3 & \cellcolor{reggain}{-0.0} \\
\midrule
Qwen 3.5-397B-FP8 & L & 72.8 & \cellcolor{lowgain}73.0 & \cellcolor{highgain}79.9 & 58.2 & \cellcolor{highgain}61.2 & \cellcolor{highgain}62.6 & 65.7 & \cellcolor{reggain}61.9 & \cellcolor{reggain}62.4 & 69.4 & \cellcolor{reggain}67.9 & \cellcolor{reggain}66.9 & 44.1 & \cellcolor{highgain}48.6 & \cellcolor{highgain}52.5 & 57.5 & \cellcolor{lowgain}58.3 & \cellcolor{lowgain}58.9 & 62.5 & \cellcolor{lowgain}62.9 & \cellcolor{medgain}65.1 & \cellcolor{medgain}{+2.6} \\
Qwen 3.5-122B-A10B & L & 73.9 & \cellcolor{reggain}72.2 & \cellcolor{reggain}73.7 & 59.9 & \cellcolor{reggain}57.5 & \cellcolor{highgain}64.4 & 59.4 & \cellcolor{lowgain}59.6 & \cellcolor{reggain}56.4 & 63.9 & \cellcolor{highgain}67.0 & \cellcolor{highgain}69.8 & 42.4 & \cellcolor{highgain}52.6 & \cellcolor{highgain}53.8 & 56.9 & \cellcolor{lowgain}57.3 & \cellcolor{highgain}60.1 & 60.9 & \cellcolor{lowgain}61.8 & \cellcolor{medgain}63.8 & \cellcolor{medgain}{+3.0} \\
Qwen 3.5-35B-A3B & M & 57.9 & \cellcolor{highgain}62.4 & \cellcolor{highgain}63.0 & 45.4 & \cellcolor{highgain}49.7 & \cellcolor{medgain}48.1 & 50.2 & \cellcolor{highgain}55.5 & \cellcolor{highgain}54.8 & 51.1 & \cellcolor{highgain}56.5 & \cellcolor{highgain}58.9 & 38.0 & \cellcolor{medgain}40.6 & \cellcolor{highgain}44.9 & 57.8 & \cellcolor{reggain}48.0 & \cellcolor{reggain}49.4 & 51.0 & \cellcolor{medgain}53.1 & \cellcolor{medgain}53.9 & \cellcolor{medgain}{+2.9} \\
Qwen 3.5-9B & S & 37.1 & \cellcolor{reggain}35.1 & \cellcolor{medgain}39.5 & 30.2 & \cellcolor{medgain}32.3 & \cellcolor{highgain}34.5 & 38.0 & \cellcolor{reggain}35.0 & \cellcolor{reggain}32.3 & 35.0 & \cellcolor{highgain}45.7 & \cellcolor{highgain}41.1 & 26.8 & \cellcolor{highgain}33.4 & \cellcolor{highgain}35.9 & 42.3 & \cellcolor{reggain}38.8 & \cellcolor{reggain}35.4 & 35.4 & \cellcolor{lowgain}36.5 & \cellcolor{lowgain}36.5 & \cellcolor{lowgain}{+1.2} \\
\midrule
GPT-oss-120B & L & 77.1 & \cellcolor{lowgain}77.4 & \cellcolor{lowgain}77.2 & 71.0 & \cellcolor{lowgain}71.1 & \cellcolor{lowgain}71.9 & 59.2 & \cellcolor{highgain}64.1 & \cellcolor{highgain}64.4 & 75.2 & \cellcolor{reggain}71.8 & \cellcolor{reggain}73.8 & 52.1 & \cellcolor{highgain}59.4 & \cellcolor{highgain}57.4 & 58.9 & \cellcolor{reggain}58.7 & \cellcolor{lowgain}59.6 & 66.6 & \cellcolor{lowgain}67.9 & \cellcolor{medgain}68.2 & \cellcolor{medgain}{+1.6} \\
GPT-oss-20B & M & 62.7 & \cellcolor{reggain}59.1 & \cellcolor{lowgain}63.6 & 59.2 & \cellcolor{reggain}56.5 & \cellcolor{reggain}54.9 & 56.5 & \cellcolor{reggain}56.4 & \cellcolor{reggain}53.3 & 56.9 & \cellcolor{reggain}53.6 & \cellcolor{lowgain}57.8 & 43.1 & \cellcolor{highgain}49.0 & \cellcolor{lowgain}44.0 & 52.0 & \cellcolor{highgain}55.6 & \cellcolor{lowgain}53.4 & 56.1 & \cellcolor{reggain}55.6 & \cellcolor{reggain}55.4 & \cellcolor{reggain}{-0.4} \\
\bottomrule
\end{tabular}
\caption{Static benchmark performance on \benchname{}: average (M2,\%) per role under three skill conditions ($\varnothing$ = no skill, H = handcrafted, G = LLM-generated). Per-task pass rate is the mean over attempts of $\mathrm{tests\_passed}/\mathrm{task\_total}$; per-role and aggregate values are unweighted means across tasks. Aggregate columns show the overall pass rate and the best delta between no-skill and any skill condition. Colors mark the gain from skills: \colorbox{best}{best}, \colorbox{highgain}{>3}, \colorbox{medgain}{1.5..3}, \colorbox{lowgain}{0..1.5}, \colorbox{reggain}{<0}.}
\label{tab:perfect-rate-test-subset}
\end{table*}

\newpage
\section{Single Refinement Results}
\label{app:e1-results}
\begin{table*}[ht]
\centering
\scriptsize
\setlength{\tabcolsep}{2pt}
\begin{tabular}{@{}ll|cc|cc|cc|cc|cc|cc|ccc@{}}
\toprule
& & \multicolumn{2}{c|}{\textbf{DE}} & \multicolumn{2}{c|}{\textbf{DS}} & \multicolumn{2}{c|}{\textbf{GenAI}} & \multicolumn{2}{c|}{\textbf{Infra}} & \multicolumn{2}{c|}{\textbf{PM}} & \multicolumn{2}{c|}{\textbf{SWE}} & \multicolumn{3}{c}{\textbf{Aggregate}} \\
\textbf{Model} & \textbf{Size} & H$_{\text{pre}}$ & H$_{\text{post}}$ & H$_{\text{pre}}$ & H$_{\text{post}}$ & H$_{\text{pre}}$ & H$_{\text{post}}$ & H$_{\text{pre}}$ & H$_{\text{post}}$ & H$_{\text{pre}}$ & H$_{\text{post}}$ & H$_{\text{pre}}$ & H$_{\text{post}}$ & Avg$_{\text{pre}}$ & Avg$_{\text{post}}$ & $\Delta$ \\
\midrule
Qwen 3.5-9B & S & 22.1 & \cellcolor{medgain}23.6 & 11.5 & \cellcolor{reggain}10.0 & 15.3 & \cellcolor{highgain}19.3 & 13.8 & \cellcolor{best}20.8 & 10.8 & \cellcolor{highgain}15.4 & 13.2 & \cellcolor{best}19.3 & 14.4 & \cellcolor{highgain}18.1 & +3.7 \\
Qwen 3.5-35B-A3B & M & 43.3 & \cellcolor{best}52.8 & 25.5 & \cellcolor{best}36.7 & 31.6 & \cellcolor{highgain}36.8 & 23.8 & \cellcolor{best}31.2 & 13.1 & \cellcolor{reggain}10.3 & 20.0 & \cellcolor{best}29.8 & 26.2 & \cellcolor{best}32.9 & +6.7 \\
Qwen 3.5-122B-A10B & L & 52.9 & \cellcolor{medgain}54.2 & 33.5 & \cellcolor{best}40.0 & 34.2 & \cellcolor{highgain}38.6 & 45.0 & \cellcolor{best}52.1 & 19.2 & \cellcolor{medgain}20.5 & 29.5 & \cellcolor{highgain}31.6 & 35.7 & \cellcolor{highgain}39.5 & +3.8 \\
\bottomrule
\end{tabular}
\caption{Refined handcrafted (H) catalogue on \benchname{}. H$_{\text{pre}}$ = canonical handcrafted skills; H$_{\text{post}}$ = single pass Codex-refined skills. Average (M2,\%) per role metric on the 111 task IDs present in both pre and post runs. Cell colours on $H_{\text{post}}$ = $\Delta$: \colorbox{best}{$\geq\!+6$}, \colorbox{highgain}{$+2\!..\!+6$}, \colorbox{medgain}{$0\!..\!+2$}, \colorbox{reggain}{$<\!0$}.}
\label{tab:e1-refine-h}
\end{table*}

\section{Example Skills}
\label{app:examples}

\begin{figure*}[ht]
\begin{lstlisting}[style=promptstyle,
  caption={\textit{Handcrafted body for the \texttt{sql} skill. One narrow task: parse an unknown SQLite database into structured JSON. Excerpt of the first two steps; the full body has four steps and a debugging-tips section.}},
  label={lst:sql-h}]
---
name: sqlite-map-parser
description: Parse SQLite databases into structured JSON data. Use when exploring unknown database
  schemas, understanding table relationships, and extracting map data as JSON.
---

# SQLite to Structured JSON

Parse SQLite databases by exploring schemas first, then extracting data into structured JSON.

## Step 1: Explore the Schema

  -- List all tables
  SELECT name FROM sqlite_master WHERE type='table';

  -- Inspect table schema
  PRAGMA table_info(TableName);     -- (cid, name, type, notnull, dflt, pk)
  SELECT sql FROM sqlite_master WHERE name='TableName';

  -- Find primary / unique keys
  PRAGMA index_list(TableName);

## Step 2: Understand Relationships

  PRAGMA foreign_key_list(TableName);
  -- Common pattern: tables share an ID column (LEFT JOIN by ID)
  -- Spatial keys:  x = id % width;  y = id // width

[... Step 3: sqlite3 + json extraction loop (row_factory=Row, dict(row), nested joins).
     Step 4: map vs array output shaping based on natural keys.
     Debugging tips for missing tables / null columns.  (Sections omitted for space.) ...]
\end{lstlisting}
\end{figure*}

\begin{figure*}[ht]
\begin{lstlisting}[style=promptstyle,
  caption={\textit{LLM-generated body for the \texttt{sql} skill. Broad reference spanning seven sections (language constructs, optimization, access libraries, worked examples, access patterns). Excerpt of the front matter and the first dispatch table.}},
  label={lst:sql-g}]
---
name: sql
description: "Reference for writing and tuning SQL on tabular sources: SELECT/INSERT/UPDATE/DELETE,
  INNER/LEFT/RIGHT/FULL/SEMI/ANTI joins, window functions (ROW_NUMBER, RANK, LAG/LEAD, running
  aggregates), recursive and non-recursive CTEs, EXPLAIN/EXPLAIN ANALYZE, index design, and three
  Python access layers (DuckDB, SQLAlchemy, psycopg2). Includes worked feature-extraction and
  slow-query-diagnosis examples."
metadata:
  dependencies:
    - duckdb
    - sqlalchemy
    - psycopg2-binary
---

# SQL Reference

Three steps: identify the question shape, pick the SQL technique (Section 1), pick the access
library (Section 3). Optimization is mechanical once you have a plan (Section 2).

## 1. Question to SQL technique

| Question                | Technique                  | Skeleton                                          |
|-------------------------|----------------------------|---------------------------------------------------|
| Filter / top-N overall  | WHERE + ORDER BY LIMIT     | WHERE col=? ORDER BY s DESC LIMIT 10              |
| Top-N per group         | ROW_NUMBER() window        | ROW_NUMBER() OVER (PARTITION BY g ORDER BY s)     |
| Rank with ties          | RANK() / DENSE_RANK()      | RANK() OVER (ORDER BY s DESC)                     |
| Running / moving sum    | SUM() OVER with frame      | SUM(x) OVER (PARTITION BY g ORDER BY t)           |
| Previous / next row     | LAG / LEAD                 | LAG(x,1) OVER (PARTITION BY g ORDER BY t)         |
| Pivot rows to columns   | conditional aggregation    | SUM(CASE WHEN k='A' THEN v ELSE 0 END)            |
| Composable subquery     | non-recursive WITH         | WITH a AS (...), b AS (...) SELECT ...            |
| Rows in A not in B      | LEFT JOIN ... WHERE NULL   | LEFT JOIN b WHERE b.k IS NULL  (or EXCEPT)        |

## 2. Optimization checklist  [EXPLAIN ANALYZE, indexes on filter/join cols, drop SELECT *, ...]
## 3. Library dispatch        [DuckDB / sqlite3 / SQLAlchemy / psycopg2]
## 4. Connection patterns     [parameterized snippets per library]
## 5. Worked example: feature extraction over Parquet (DuckDB + CTE + window aggregate)
## 6. Worked example: diagnose a slow query via EXPLAIN ANALYZE
## 7. Efficient access patterns  [B-tree composite, columnar store, partition by ts]

[... Pitfalls section omitted ...]
\end{lstlisting}
\end{figure*}

Every paper-target skill in \benchname{} ships with two static bodies: a \emph{handcrafted} body (H), authored against a concrete user-facing scenario, and an \emph{LLM-generated} body (G), drafted by a frontier model as a broad procedural reference. We illustrate the contrast on the \texttt{sql} skill in Listings~\ref{lst:sql-h} and~\ref{lst:sql-g}. The handcrafted body (Listing~\ref{lst:sql-h}) solves one narrow task --- parsing an unknown SQLite database into structured JSON; the LLM-generated body (Listing~\ref{lst:sql-g}) is a seven-section cheatsheet that covers join shapes, window functions, optimization, and three Python access layers.
\end{document}